\documentclass[lettersize,journal]{IEEEtran}
\usepackage{amsmath,amsfonts}
\usepackage{algorithm}
\usepackage{array}
\usepackage[caption=false,font=normalsize,labelfont=sf,textfont=sf]{subfig}
\usepackage{textcomp}
\usepackage{stfloats}
\usepackage{url}

\usepackage{verbatim}
\usepackage{graphicx}
\usepackage{cite}

\usepackage{algpseudocode}
\usepackage{amsfonts}
\usepackage{enumitem}
\usepackage{multirow}
\usepackage{titlesec}
\usepackage{color}
\usepackage{colortbl}

\begin{document}

\title{Beyond Prediction: On-street Parking Recommendation using Heterogeneous Graph-based List-wise Ranking}

\author{
\thanks{The research described in this study was supported by the Smart Traffic Fund of the Transport Department of the Hong Kong Special Administrative Region, China (ref. no.: PSRI/07/2108/PR) and a grant from the Research Institute for Sustainable Urban Development (RISUD) at the Hong Kong Polytechnic University (project no. P0038288). The authors thank the Transport Department of the Government of the Hong Kong Special Administrative Region for providing the on-street parking data. The contents of this paper reflect the views of the authors, who are responsible for the facts and the accuracy of the information presented herein. (Corresponding author: Wei Ma).}
Hanyu Sun\thanks{H. Sun is  with the  Department  of  Civil and  Environmental  Engineering, The  Hong  Kong  Polytechnic  University,  Hong  Kong  SAR,  China (E-mail:hanyu.sun@polyu.edu.hk).}, Xiao Huang\thanks{X. Huang is  with  the Department of Computing, The  Hong  Kong  Polytechnic  University,  Hong  Kong  SAR,  China (E-mail:xiaohuang@comp.polyu.edu.hk)}, Wei Ma,~\IEEEmembership{Member, IEEE}\thanks{W. Ma is with the Department of Civil and Environmental Engineering, The Hong Kong Polytechnic University, Hong Kong SAR, China; Research Institute for Sustainable Urban Development, The Hong Kong Polytechnic University, Hong Kong SAR, China (E-mail: wei.w.ma@polyu.edu.hk).}
}

%


\markboth{Manuscript Submitted to IEEE Transactions on Intelligent Transportation Systems, 2023}
{}

\maketitle

\begin{abstract}
To provide real-time parking information, existing studies focus on predicting parking availability, which seems an indirect approach to saving drivers' cruising time.
In this paper, we first time propose an on-street parking recommendation (OPR) task to directly recommend parking spaces for a driver.
To this end, a learn-to-rank (LTR) based OPR model called OPR-LTR is built. 
Specifically, parking recommendation is closely related to the ``turnover events'' (state switching between occupied and vacant) of each parking space, and hence we design a highly efficient heterogeneous graph called ESGraph to represent historical and real-time meters' turnover events as well as geographical relations; afterward, a convolution-based event-then-graph network is used to aggregate and update representations of the heterogeneous graph.  A ranking model is further utilized to learn a score function that helps recommend a list of ranked parking spots for a specific on-street parking query. The method is verified using the on-street parking meter data in Hong Kong and San Francisco. By comparing with the other two types of methods: prediction-only and prediction-then-recommendation, the proposed direct-recommendation method achieves satisfactory performance in different metrics. Extensive experiments also demonstrate that the proposed ESGraph and the recommendation model are more efficient in terms of computational efficiency as well as saving drivers' on-street parking time.
\end{abstract}


\section{Introduction}
\IEEEPARstart{W}{ith} much time used in searching for on-street parking spaces in megacities, traffic jams are caused, extra energy is wasted, environments are polluted and at last huge economic losses are incurred. A recent survey shows that Americans spend an average of 17 hours per year searching for parking, resulting in a cost of \$345 per driver
\cite{USATraffic}. As most waste is caused by time consumption in finding a parking space, we thus turn to develop a method that aims to save drivers' time in cruising for parking.

Recently, smart parking projects have utilized sensing technology to help drivers find vacant parking spaces in order to reduce the cruising time\cite{schneble2021statistical}. Existing sensing technology uses either identification technologies like stationary sensors or communication technologies of vehicular information like crowd-sourcing-based methods to monitor the status of parking spaces.
In some smart parking projects, on-street parking meters equipped with stationary sensors are used to record the parking status in a fine-grained time level. Though large-scale sensor network deployment is costly, these meters are capable of collecting real-time parking availability information and will bring much more economic benefits in the future.
On another side, crowd-sourcing-based methods like those studied by Chen et al.\cite{zheng2019analytical} and Jim et al.\cite{cherian2016parkgauge}, though cost-efficient, suffer the issues of information accuracy, participation rates, and free-rides\cite{chen2012crowdsourcing}. Comparatively, meters with stationary sensor data have natural advantages in analyzing city-wide on-street parking patterns thus utilized in our project.
Recently, the Transport Department in Hong Kong has completed the utilization of new parking meters with millimeter-wave radar installed to detect real-time occupancy of parking spaces at a citywide scale\cite{wu2022clustering}. Similar equipment is also utilized in many megacities, such as SFpark in San Francisco and LA Express Park in Los Angeles.

\begin{figure}[!t]
  \centering
  \includegraphics[width=\linewidth]{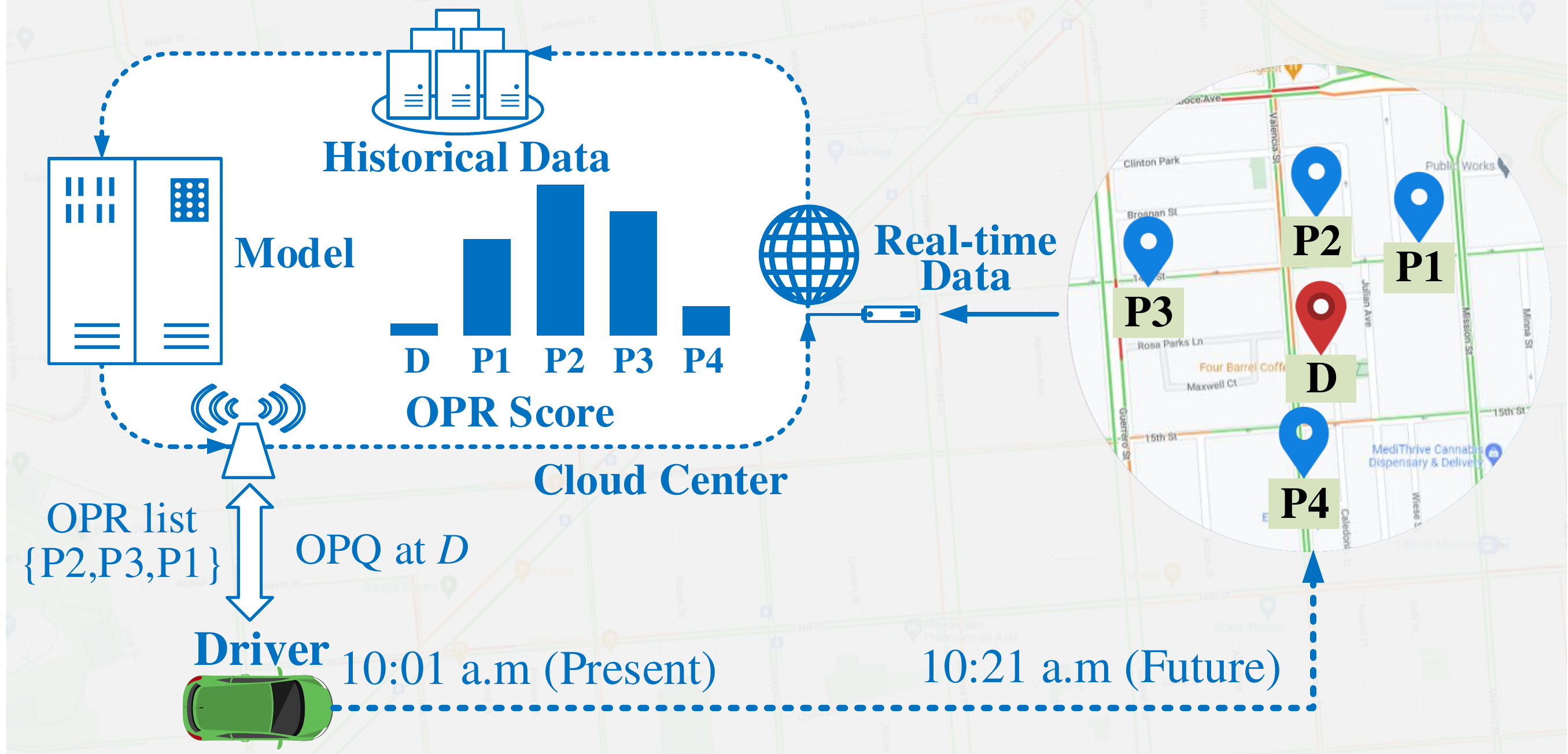}
  \caption{An OPR example. A driver sends a parking request at destination \textit{D} at 10:01, then receives a recommended list from an OPR model with ranked results of the destination.}
  \label{img1}
\end{figure}

The latest on-street parking model makes use of the meter data and focuses on on-street parking availability prediction \cite{7112165,9905231,9997228, 9507392, liu2020graphsage}, Zhao et al. \cite{10.1109/TITS.2021.3067675} built a MePark model to predict real-time city-wide on-street parking availability using meter data and achieved better prediction accuracy than baseline models. Though an accurate prediction of parking availability is of significant importance, it is difficult to guarantee the precision for the inaccuracy and complex spatiotemporal correlation of data as well as the uncertainty and bias of the model \cite{rodrigues2020beyond}. Besides, only predicting parking availability is not practically interesting for smart parking application because it neither meets the ultimate purpose of saving time consumption nor help drivers make a final decision on where to park. Imaging the following two scenarios:

\begin{itemize}
    \item A user receives a predicted result that all the parking spaces are occupied around the intended destination, and this happens frequently during rush hours in megacities.
    \item A user arrives at a predicted available destination but found it is occupied due to either the predicting error or parking competition.
\end{itemize}

Both scenarios are out of the range of prediction-only models and will then cost drivers extra time to find a suitable parking space. To handle these issues, a straightforward solution is to make an extra effort to guide drivers based on the predicting model's result like tried by Liu \cite{liu2018street} and Zhao \cite{zhao2020d2park}; however, their guidance completely depends on the results of predicting model. Besides, the two-step approach has the problem of incoherence and requiring extra resources (both space and time) thus causing bad guidance and time wasting. Given the above, we propose a one-step approach that combines the prediction and allocation tasks into one task with the aim of providing the driver with a list of recommended on-street parking spaces, furthermore, we rank the recommended list to a permutation with better candidates ranked forwardly, we then called the task on-street parking recommendation (OPR), as illustrated in Fig.~\ref{img1}. 

Turnover events, which describe the state switching of parking space between vacant and occupied states, are critical features for parking recommendation, and this has not been explored in previous studies. However, we notice that the parking time series data are often recorded in a fixed interval, and it is challenging to consider or analyze the turnover events in the conventional spatiotemporal graph \cite{al2020characteristics}. 
This motivates us to develop novel graph construction and representation methods for parking data.

As a specific type of spatiotemporal (ST) data, citywide on-street parking data \cite{pu2017evaluation, 9061155} varies in both spatial regions and temporal domains that are usually represented by a graph and time sequences separately. Spatial and temporal features can be embedded in different orders \cite{gao2022equivalence}, referring to as the time-and-graph and time-then-graph, respectively. Recently, heterogeneous graphs have been used to combine different types of relevant data and achieved good performance \cite{10.1145/3331184.3331273,liu2020heterogeneous,zhao2022space4hgnn}. This inspires us to integrate on-street spatiotemporal data in the Event-Spatial Heterogeneous Graph (ESGraph), and we further develop the concept of event-then-graph. Importantly, the turnover event of parking spaces can be represented using the path concatenation in the heterogeneous graph.

To summarize, beyond only predicting on-street parking availability, we recommend a list of ranked parking spaces so as to help drivers make parking decisions to save time in cruising for parking. To this end, we first define the OPR task formally in the aspect of recommendation. Then a heterogeneous graph is designed to better express the original data in which turnover event information is included. Following this, an event-then-graph convolutional layer is designed to better aggregate ESGraph features in an efficient way. At last, a score function is learned to extract interdependencies for the final OPR task.     
We validate the proposed model using real-world data in Hong Kong and San Francisco. 
In comparison, we consider three types of models: prediction-only models; prediction-then-recommendation models; direct-recommendation models (e.g., our proposed model), then evaluate their corresponding performance in recommendation metrics like NDCG and MAP as well as saving cruising time for drivers. Our main contributions are as follows:
\begin{itemize}[leftmargin=*]
\item We first time propose an on-street parking recommendation (OPR) task to directly provide practical and executable suggestions to drivers for each request with a specific destination. 
\item  We design an Event-Spatial Heterogeneous Graph (ESGraph) to represent citywide on-street parking data and prove its advantages in representation power and space\&time complexity;
\item An event-then-graph convolutional layer is developed to aggregate ESGraph representation for the event interdependency learning of OPR.
\item The experimental results validate that our proposed model is more straightforward and effective than prediction-based models, and it outperforms other state-of-art recommendation models.
\end{itemize}

\section{Related Work}

We discuss parking prediction models and conventional recommendation models in this section.

\subsection{Data-driven Parking Prediction Models}
Although prediction is not the focus of this study, the study will compare with standard prediction models and demonstrate the merits of recommendation models. 
Firstly, data-driven Parking Prediction utilizes statistical models trained by historical and real-time observations to predict aggregated parking occupancies in a mesoscopic manner \cite{yang2019deep}. Latest parking prediction models used GCN and RNN to extract spatiotemporal features like Yang \cite{yang2019deep}, MePark \cite{zhao2020d2park}
and Zhang \cite{zhang2020semi}. Some other state-of-art spatiotemporal data processing models are also worth mentioning and trying, such as Diffusion Convolutional Recurrent Neural Network (DCRNN) \cite{li2017diffusion} used for traffic forecasting and Spatial-Temporal Position-Aware Graph Convolution Networks (STPGCN) for traffic flow forecasting\cite{9945663}.

\subsection{Learn-To-Rank Models}

Learning to rank (LTR) automatically constructs a ranking model using training data, such that the model can sort new items according to their degrees of relevance \cite{10.1145/1835449.1835676}. Among all LTR models, list-wise approaches like ListNet \cite{cao2007learning}, ListMLE \cite{xia2008listwise} and ListMAP \cite{keshvari2022listmap} were thought to be better than point-wise and pair-wise methods as they extracted inter-item dependencies in a loss function level \cite{datta2022pointwise, wang2022meta}. Recently, methods like Seq2Slate \cite{bello2018seq2slate} considered inter-item dependencies in a score function's level and achieved good performance. State-of-art LTR models like ApproxNDCG \cite{qin2010general} and NeuralNDCG \cite{pobrotyn2021neuralndcg} also tried to optimize ranking metrics directly and achieved better performances than some baseline models. A typical ranking data set with $N$ lists can be denoted as $\{(\boldsymbol{q},\boldsymbol{X},\boldsymbol{y})\}_1^N$. In each list, $\boldsymbol{q}$ is the query feature vector; $\boldsymbol{X}=\{\boldsymbol{x_i}\}_{i=1}^l$ is a set of $l$ items, each represented by a feature vector $\boldsymbol{x_i}$; $\boldsymbol{y}=\{y_i\}_{i=1}^l$ is each item's relevant label \cite{zhuang2021interpretable}. LTR models are frequently used in information retrieval, to the best of our knowledge, no LTR model has been explored in an on-street parking problem.

\section{Problem Formulation}
\label{sec:formulation}
In this section, we formulate the OPR task in the LTR setting, and we first present some definitions regarding the LTR setting.

\noindent {\bfseries Definition 3.1:} \textit{On-street Parking Query (OPQ).} Given an on-street parking query (request) for a destination $d$ at time $t$, using an n-dimension feature vector $\boldsymbol{Q_{d,t}} \in \mathbb{R}^n$ to represent this request, where $d$ is a vertex on the graph representing the geographical information of the destination, $t$ is the request time, and $\boldsymbol{Q_{d,t}}$ contains the details of the requests ({\em e.g.}, geo-coordinates, requirements, {\em etc}). 

\noindent {\bfseries Definition 3.2:} \textit{Relevant Items of OPR (RIO).} Given an OPQ with $\boldsymbol{Q_{d,t}}$, its $RIO$, $\boldsymbol{X_{t^\prime}}=\{\boldsymbol{X_{j,t^\prime}}\}_{j=1}^l$, is a list of $l$ items (parking space candidates) at a future time $t^\prime$. Each item $\boldsymbol{X_{j,t^\prime}}\in \boldsymbol{X_{t^\prime}}$ is represented by a $m$-dimension feature vector $(x_1,x_2,...,x_m)$ as well as a relevant label $y_j(\boldsymbol{y}=\{y_j\}_{j=1}^m)$. 

\begin{figure}[!t]
  \centering
  \includegraphics[width=\linewidth]{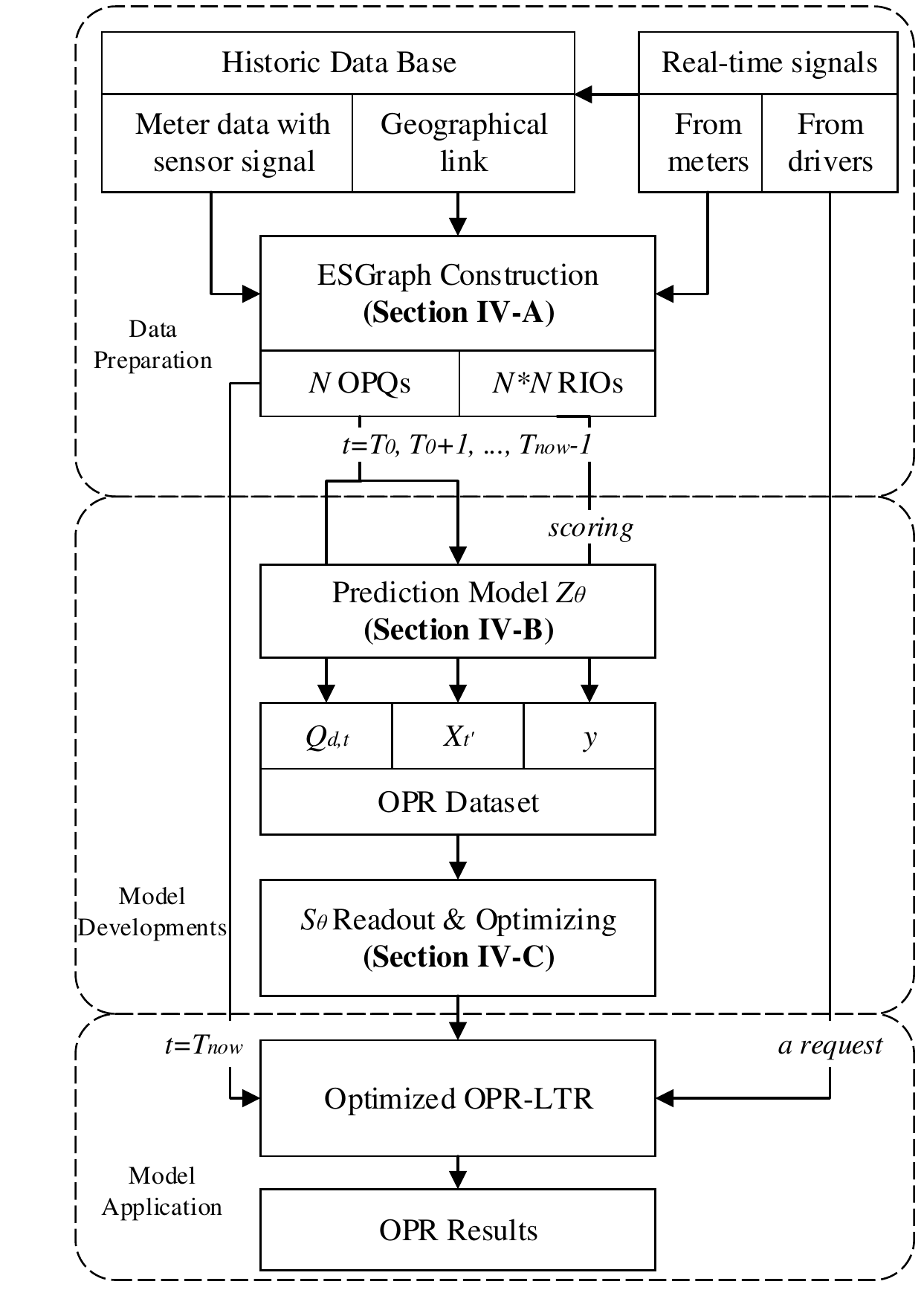}
  \caption{Flow chat of the proposed OPR framework.}
  \label{flowchat}
\end{figure}

Based on the above definitions, we then can formulate the OPR problem. Let OPQ and RIO represent the on-street parking spaces, then all possible on-street parking spaces form a full set $\mathcal{P}$ in which $\lVert \mathcal{P}\rVert
=N$ is the total number of these spaces. Afterwards we have the OPR data set $\mathcal{D}=\{(\boldsymbol{Q_{d,t}},\boldsymbol{X_{t^\prime}},\boldsymbol{y}):\forall d\in \mathcal{P}\}_{t=1}^T$. We aims to learn a relevant score function $S_\theta$ with $\theta$ as its parameter from data set $\mathcal{D}$, such that for an arbitrary OPQ, $S_\theta$ takes into $(\boldsymbol{Q_{d,t}},\boldsymbol{X_{t^\prime}})$ as input and predict a list of relevant scores $\boldsymbol{\hat{y}}$:
\setlength\abovedisplayskip{3pt}
\setlength\belowdisplayskip{3pt}
\begin{equation}
  \boldsymbol{\hat{y}}=S_\theta(\boldsymbol{Q_{d,t}},\boldsymbol{X_{t^\prime}}).
\label{con:F}
\end{equation}
Note that different from the easy availability of $\boldsymbol{Q_{d,t}}$ feature, which can get from OPQ's real-time and previous data, it is hard to extract $\boldsymbol{X_{t^\prime}}$ feature for it happens in a future time. In this work, we used a model $Z_\theta$ with $\theta$ as its parameter to predict $\boldsymbol{X_{t^\prime}}$ along with multiple already available features till time $t$:
\begin{equation}
  \boldsymbol{\hat{X}_{t^\prime}}=Z_\theta(\boldsymbol{Q_{d,\leq t}},\boldsymbol{X_{\leq t}}).
\end{equation}
Then the input space of the model includes previous and present parking states $\boldsymbol{Q_{d,\leq t}},\boldsymbol{X_{\leq t}}$, especially the previous features will be extracted in a turnover event-based manner (which will be discussed in section \ref{sec:method}). The output space of the model is compared to the ground true label $\boldsymbol{y}$. The hypothesis space of the model is represented by the relevant score function $F$. Then we can treat the goal of the OPR-LTR model as  minimizing the average distance between the relevant label and the predicted relevant score for all the $N$ queries by
\begin{equation}
  \frac{1}{N}\sum\limits_{i=1}^{N}{L(\boldsymbol{\hat{y}},\boldsymbol{y})}.
\label{L}
\end{equation}
Based on the predicted list of the relevant scores, we can make a top $n$ recommendation to the driver. An overview of the entire OPR framework is presented in Fig. \ref{flowchat}.

\section{Methodology}
\label{sec:method}
\begin{figure*}[!t]
  \centering
  \includegraphics[width=1\linewidth]{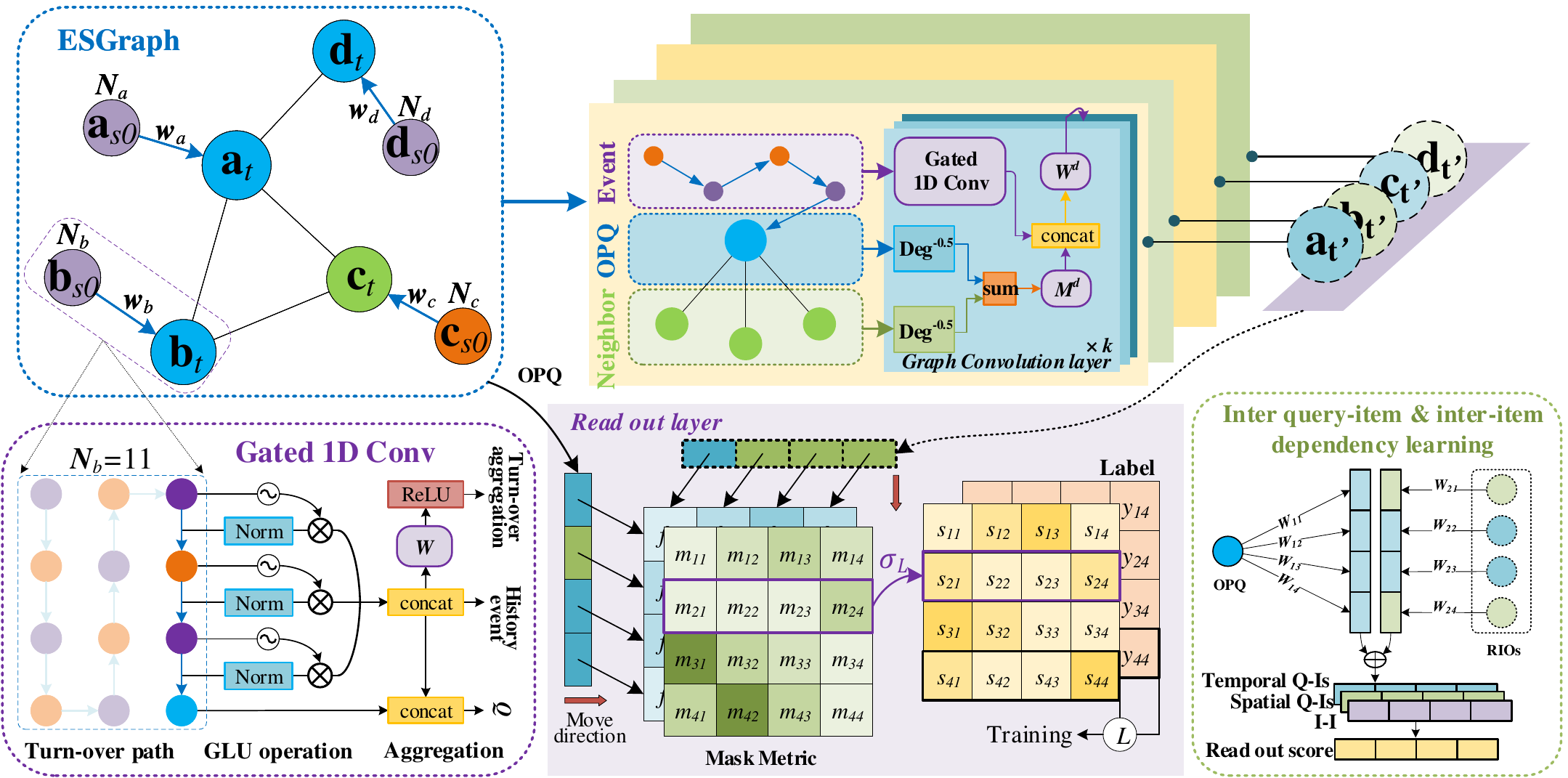}
  \caption{Overview of the model architecture of proposed OPR-LTR. Following ESGraph, the proposed OPR-LTR model runs according to the following workﬂow: the data flow of ESGraph is firstly inputted into an Event-then-graph Convolutional Layer, in which a Gated 1D Conv layer is employed to aggregate the N-length turn-over event path of each node (OPR), then together with this node and all its adjacent neighbors, a graph convolutional network is used to combine, update, and then output a final representation of this layer, as represented by $a_{t'},b_{t'},c_{t'},d_{t'}$ in the figure. Afterward, these final outputs (act as RIOs) and ESGraph's nodes (act as OPQs) are input into the Readout Layer, and this layer uses a learning-based score function to extract features of Q-Is and I-Is to derive score values. In the picture of Inter query-item and inter-item dependency learning, each OPQ corresponds to a list of RIOS will derive a list of predicted scores through the learnable score function. Finally, a matrix combining all lists will be compared with the label to optimize the entire model.}
  \label{img3}
\end{figure*}

In this section, we build the ORP-LTR model in detail. We first design a heterogeneous graph to represent the historical turnover events and spatiotemporal features for parking vacancy; then a CNN-based event-then-graph network is built to aggregate heterogeneous features and the network output is further forwarded to the final task-based layer for producing the relevant scoring lists for candidate parking spaces. An overview is shown in Fig.~\ref{img3}.  

\subsection{Graph Construction}

\begin{figure}[!t]
  \centering
  \includegraphics[width=\linewidth]{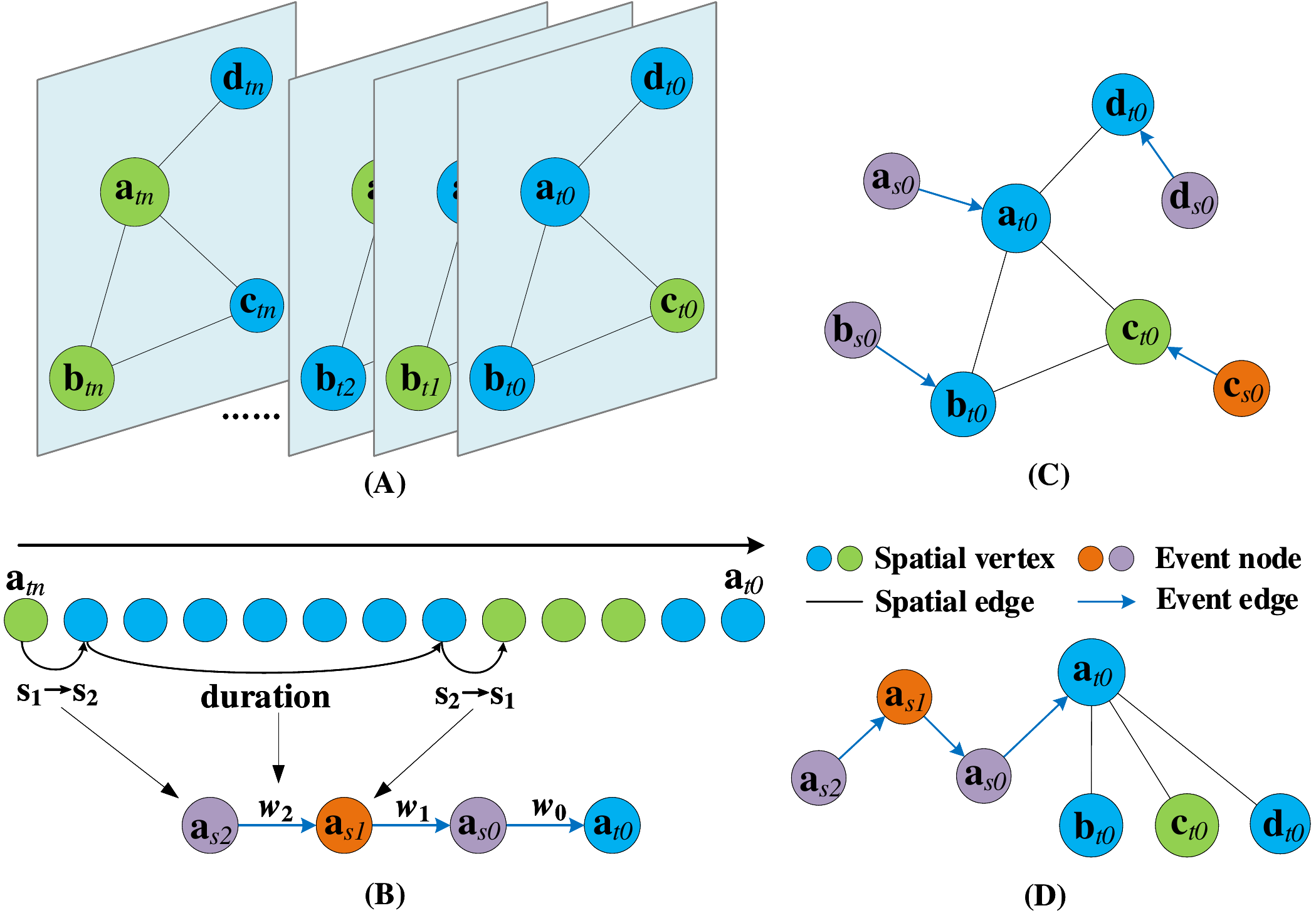}
  \caption{Converting an STGraph to an ESGraph. (A) A sequence of spatial graphs from a previous time $t_n$ to a present time $t_0$ with fixed time interval. (B) Edge-contraction in one temporal path of STGraph. (C). The resulting event-spatial heterogeneous graph with 1 turn-over event. (D) One spatial vertex with its 1-D spatial adjacent and 3 turn-over events.}
  \label{img2}
\end{figure}

Based on the original graph (which is referred to as STGraph), we build our proposed heterogeneous graph. An overview of the procedures for constructing the heterogeneous graph is shown in Fig.~\ref {img2}. 

\noindent {\bfseries Definition 4.1} \textit{SGraph.} 
A spatial graph (SGraph) $(V_s, E_s)$ represents the  correlation of geographical vertices, where $V_s$ and $E_s$ are the location-typed vertex set and adjacent-based edge set, $v=\lVert V_s \rVert$ is the total number of vertices in a graph.

\noindent {\bfseries Definition 4.2} \textit{STGraph.}  
A spatial-temporal Graph (STGraph) of length $T$ can be seen as a sequence of SGraphs over $T$ consecutive times, $\{(V_{s,t}, E_{s,t}):t=1,\cdots, T\}$, where $t$ is the time index of each SGraph and the time interval is fixed.   

\noindent {\bfseries Definition 4.3} \textit{PGraph.}
A path graph  (PGraph) is a graph whose all $T$ nodes are listed in order of $\{n_1, n_2, ..., n_{T}\}$ and edges are  $(n_i,n_{i+1})$ where $i=1,2,..., T-1$ \cite{diestel2005graph}.


Note that in the STGraph, the $T-$length temporal sequence of $i^{th}$ vertex can be treated as a PGraph $(V_{i,t}, E_{i,t})$ in which node $V_{i,t}$ is listed in order of $\{n_{i,1}, n_{i,2}, ..., n_{i,T}\}$ and  $E_t=\{(n_{i,t},n_{i,t+1}): t=1,2,...,n-1\}$ is the edge. A STGraph can then be treated as a set of $v$ PGraphs $\{(V_{i,t}, E_{i,t}):i=1,2,...,v\}$ or short for temporal path $(V_t,E_t)$.

\subsubsection{\textit{Path Contraction for Turn-over Event.}}

In a STGraph that represents on-street parking meter data, as shown in Fig.~\ref{img2}, all its temporal paths are represented in fixed time intervals in which we can use an \textit{edge-contraction} technology to bring in turnover events. A graph that conducts edge-contraction will have its contracted edges $e$ removed and incident nodes of $e$ merged into a new node \cite{gross2018graph}, the resulting induced graph is written as $G\backslash e$. Specifically in a temporal path $(V_t, E_t)$, starting from $t=1$, edges $(v_t,v_{t+1})\in E_t$ whose incident nodes owning the same contribution will be removed, $v_{t+1}$ is then merged into $v_t$ and at each time of contracting, next edge $(v_{t+1},v_{t+2})\in E_t$ will be weighted more. Then our path contraction process can be represented by a map function $f: (V_t, E_t)\rightarrow (U, \mathcal{E}, W_{\mathcal{E}})$ such that:
\begin{equation}
\begin{array}{l}
U=V_t\backslash V_{t^\prime}, \\
\mathcal{E}=E_t\backslash \{(n_i,n_{i +1}):n_i\in V_{t^\prime} \}_{i=1}^{\lVert V_{t^\prime} \rVert}, \\
W_{\mathcal{E}}=\{\textit{coun}(u_j):u_j\in U \}_{j=1}^{\lVert \mathcal{E} \rVert},
\end{array}
\label{con:edge-contraction}
\end{equation}
where $V_{t^\prime}$ is made up of all merged nodes, $\lVert \cdot \rVert$ indicates the number of elements in a set, and \textit{coun($\cdot$)} computes the number of contractions that happened in a node of the temporal path. The result is a new-typed PGraph $(U, \mathcal{E}, W)$ with fewer nodes and edges than the original temporal path. Noticed that in $(U, \mathcal{E}, W)$, every edge has different contributed (stated) incident nodes, and every edge $(u_i,u_{i+1})\in U$ represents the state duration of $u_i$ which makes every $u_i$ a state switching node. In a two-state condition like on-street parking's vacant/occupied state, $(U, \mathcal{E}, W)$ can represent \textit{turnover event}. As shown Fig.~\ref {img2}(B) is an illustration of the path graph's edge contraction in a 2-feature example. We then call this PGraph \textit{turn-over event} based path. 
\subsubsection{Event-spatial Heterogeneous Graph.}
By replacing all temporal paths in a STGraph with their corresponding turnover event-based paths (short for event path), we obtain a heterogeneous graph with space-based and turnover event-based (short for event-based) vertices and edges.

\noindent {\bfseries Definition 4.4} \textit{ESGraph.} 
A Event-Spatial Heterogeneous Graph (ESGraph) $(V, E, U, \mathcal{E}, W_\varepsilon)$ has spatial typed vertices $v_i\in V$ and edges $(v_i,v_j)\in E$ as well as event-based nodes $u_{i,k}\in U$, edges $(u_{i,k},u_{i,k+1})\in \mathcal{E}$ where $(u_{i,n+1}=v_i)$ and edge weights $w_{i,k}\in W_\varepsilon$ in which $\{u_{i,k}:k=1,...,n\}\to v_i\in V$ is in a directed order from $u_{i,1}$ to $v_i$ and $n$ is the number of historic events in this path. Edge $(u_{i,n},v_i)$ is the connection between each event path and spatial graph and its weights indicating the current state duration of real-time $v_k$.

Fig.~\ref {img2} is an illustration of the process of building ESGraph using on-street parking meters' data. The original data is STGraph based with vertices' contributions of whether vacant or occupied statuses, we then transfer its temporal path to an event path by using Equation \ref{con:edge-contraction} and then merging it into an ESgraph. 
Details are available in Appendix A with Algorithm 1 and 2 the Path graph's edge contraction $\Gamma$ and on-street parking ESgraph merging. 
There are several advantages of ESGraph over STGraph summarized below: 
\begin{itemize}[leftmargin=*]
\item ESGraph data structure is more efficient in both dimensions of space and time. That is to say, ESGraph requires less space than STGraph without losing information when saving data and ESGraph need less time complexity in using data than STGraph. We proved that the amortized space and time complexity \cite{tarjan1985amortized} of ESgraph is $O(n)$ and that of STGraph is always $O(n^2)$, a detailed analysis is in Appendix B. 
\item Models based on ESGraph have smaller sizes. To be precise, when achieving comparative performances, methods based on ESGraph will use fewer parameters than methods based on STGraph. Moreover, methods using STGraph get truncated data for the reason of fixed sequential lengths while methods using ESGraph get a complete description of data for event-based representation. Detail comparations can be found in the experiments.
\item ESGraph embeds richer information. In the specific task of OPR using ESGraph with on-street parking data, $W_\varepsilon$ can provide the duration of each turnover (vacant or occupied) event, which is important when scoring the relevant degree of an RIO to an OPQ. For example, a vacant space with a longer duration will be more likely to be recommended so as to give more flexibility to the driver in case of traffic congestion.
\end{itemize}

\subsection{Event-then-graph Convolutional Layer}
\label{sec:egcl}
In this section, following the utility of ESGragh in representing on-street parking meter data, we study its feature aggregation model. Given an ESGraph $(V, E, U, \mathcal{E}, W_\varepsilon)$ containing the on-street parking data, the citywide parking meter's geographical relationship and real-time states are represented by $(V, E)$ and each meter's historical turn-over events is represented by $(U, \mathcal{E}, W_\varepsilon)$. Then we aim to build a model $Z_\theta$ with $\theta$ as its parameter, that is able to embed citywide on-street parking meters' spatial features, real-time state, and historical turn-over events together to conduct the following LTR task.

Following the schema of \textit{time-then-graph} framework \cite{gao2022equivalence}, we propose a \textit{event-then-graph} and replace the RNN-based temporal feature encoding with CNN-based turn-over event feature aggregation, then the aggregated event features together with real-time states and spatial relations are aggregated by a GNN-based network. The process is formally defined as:
\begin{equation}
\begin{array}{l}
\mathbf{H}_{i,<T}^n=\text{CNN}^n(U_{i,<T},\mathcal{E}_{i,<T},W_\varepsilon), \forall i\in V, \\
\mathbf{Z}=\text{GNN}^l(V_T, E_T, \mathbf{H}_{:,<T}^n),
\end{array}
\label{con:event-then-graph}
\end{equation}
where $\mathbf{H}$ aggregates the event features, $n$ is the number of turn-over events, $T$ represents current time, and $\mathbf{Z}$ is final representation of the event-then-graph Convolutional layer.
\subsubsection{Turnover Event Aggregation.}
We aggregate each path's $n$ turnover events' features using a gated 1-dimension convolutional architecture, which is inspired by \cite{dauphin2017language}. Firstly, a gated linear unit (GLU) was used to aggregate the event path's raw node and edge data. Specifically, raw turn-over event data of all paths can be viewed as a $n$-length $c$-path data $[X_i, W_\varepsilon]\in \mathbb{R}^{n\times c\times 2}$, which is then input into a GLU:
\begin{equation}
 X_{U\otimes\mathcal{E}}=\text{GLU}\left([X_U, W_\varepsilon]\right)=W_\varepsilon\otimes\delta(X_U),
\label{con:GLU}
\end{equation}
where $X_U$ is the contribution of $U$ and we treat $W_\varepsilon$ as the contribution of $\mathcal{E}$, $\otimes$ is the point-wise multiplication, gates $\delta(X_U)$ controls the direction of $W_\varepsilon$ like using a sign function according to whether it is vacant or occupied indicated by the value of $X_U$. The resulting $X_{U\otimes\mathcal{E}}\in\mathbb{R}^{n\times c}$ is half the size of the original input, which then needs a half number of convolutional parameters. Then we have each hidden layer as:
\begin{equation}\label{eq:aggregate}
\mathbf{H}_{:,<T}^n=\text{ReLU}(X_{U\otimes\mathcal{E}}*W+b),
\end{equation}
where $\ast$ is convolution operation, $W\in \mathbb{R}^{m\times c}$ is a learnable weight, $b\in \mathbb{R}^c$ is the bias. 
\subsubsection{Graph Convolution based Feature Updating.}
Adopting message passing-based graph neural networks \cite{dwivedi2020benchmarking}, we propose an ESGraph convolutional network to iteratively update the combined representations from event-based convolutional results and local real-time states, then output an updated representation in the final iteration:
\begin{equation}\label{eq:update}
 z_v^{(k)} = \Phi \left(\text{COMBINE}\left(g\left(\{z_{v^\prime}^{(k-1)}\}\right),h_v\right),z_v^{(k-1)}\right),
\end{equation}
where $v^\prime \in \mathcal{N}(v)$ means $v^\prime$  is the neighbor of $v$,  $h_v\in \mathbf{H}_{:,<T}^n$ is the event-based features of vertex $v$. The function $g$ aggregates local information of $v^\prime$  and function $\Phi$ updates aggregated features in $k$ iterations. $z_v^{(k)}\in \mathbf{Z}$ is the final representation of vertex $v$.

To realize the above process, we first simultaneously consider $v$ and its neighbors $v^\prime\in \mathcal{N}_v\cup \{v\}$ and represent vertex $v^\prime$ real-time local aggregation by linearly approximating its localized spectral convolution followed by a symmetric normalization \cite{kipf2016semi}. Then we use a CONCAT operation to take in events' learned features $h_v$ and serve as a residual connection \cite{he2016deep}. After which, the combined features are filtered by a convolution layer followed by an activation layer to generate the output of this iteration. The whole process will be updated $k$ times to come out with a final result. Considering vertices' contributions $X_v\in\mathbb{R}^{a\times 1},\forall v\in V$, we then have: 
\begin{equation}\small
    z_v^{(k)} = \text{ReLU}\left(\text{CONCAT}\left(g\left(z_v^{(k-1)}\right), h_v\right)*W_{v,k}+b_{v,k}\right),k>1,
\end{equation}
in which
\begin{equation}
    g\left(z_v^{(k-1)}\right)=
    \frac{1}{\sqrt{\text{deg}_v} \sqrt{\text{deg}_{v^\prime}}}\sum\limits_{v^\prime}{z_{v^\prime}^{(k-1)}}*M_{v,k},k\ge 1,
\end{equation}
Particularly, we have
\begin{equation}
z_v^{(0)} = X_{v^\prime},\forall v^\prime\in \mathcal{N}_v\cup \{v\},
\end{equation}
where $W_{v,k}\in\mathbb{R}^{{(m+c)}\times d}(k>0)$, $W^0\in \mathbb{R}^{a\times d}(k=0)$, $M_{v,k}\in \mathbb{R}^{d\times d}$ and $b_{v,k}\in \mathbb{R}^{d\times 1}$ are learnable weights and bias, $d$ is the output dimension.
while $\text{deg}_i$ and $\text{deg}_j$ are the in-degrees of node $i$ and $j$.

\subsection{Readout Layer and Objective}

In this section, we first describe $F_\theta$ in Equation \ref{con:F}, then aggregate $Z_\theta$ and $F_\theta$ for training in a supervised learning manner. Specifically in order to score RIOs, we consider the following factors in both relevant score function modeling and relevant label-creating processes:
\begin{itemize}[leftmargin=*]
\item \textit{Individual features of the query and item:} In OPR, $\boldsymbol{Q_{d,t}}$ (as defined in Definition 3.1) can be extracted from an ESGraph and $\boldsymbol{X_{t^\prime}}$ (as defined in Definition 3.2) can be modeled using the final representation of $\mathbf{Z}$ (proposed in section \ref{sec:egcl}).
\item \textit{Query-item dependency (Q-Is):} Besides the temporal Q-Is between each OPQ-RIOs pair, spatial Q-Is that indicate OPQ to RIOs distance are also of interest. Specially, we use a weighted mask layer for this spatial Q-Is \cite{he2017mask}.
\item \textit{Inter-item dependence (I-Is):} Users’ preference for a given item on a list depends on other items present in the same list \cite{pobrotyn2020context}. Given this, we adopt a learning or activation layer to take I-Is into consideration before the final readout.
\end{itemize}
       
In this study, to fulfill the completeness of RIO, all vertices (including the query vertex itself) in an ESGraph are chosen. For a $\boldsymbol{Q_{d,t}}$, we have $\boldsymbol{X_{t^\prime}}=\{\boldsymbol{X_{j,t^\prime}}\}_{j=1}^l$ in which $j\in \mathcal{P}$ is the full set and $l=N$, meaning the scorer will score a list of $N$ RIOs for each OPQ. 

\subsubsection{Learning-based Score Function.}

Based on the above discussions, we utilize a learning model to extract the above features especially to learn inter-query-item and inter-item dependencies then formulate the model $S_\theta$. First of all, we consider each query by combining the vertex's real-time state with its history turnover events (after a GLU operation): 
\begin{equation}
\boldsymbol{Q}=\text{CONCAT}(X_{U\otimes\mathcal{E}},V),
\end{equation}
where $\boldsymbol{Q}\in \mathbb{R}^{c\times (n+1)}$. Then for each $\boldsymbol{X_{j,t^\prime}}\in \boldsymbol{X_{t^\prime}}$, we use $z_j^{(k)}\in \mathbf{Z}$ from section \ref{sec:egcl} as an analog, after which we obtain the inter query-item dependency as:
\begin{equation}
f_{d,j}=\text{ReLU}(Q_d W_1+z_j^{(k)}W_2+b_1),
\end{equation}
where $Q_d\in \boldsymbol{Q}$, $W_1\in \mathbb{R}^{(n+1)\times d}$, $W_2\in \mathbb{R}^{d\times d}$ and $b_1\in \mathbb{R}^{d}$ are the learnable weights and bias, ReLU is an activation function. Because in this study $d,j\in \mathcal{P}$, which means the number of OPQ and RIO for each OPQ is equal, we then can use a matrix to fully express a citywide OPQ:
$$
M_f=\begin{pmatrix}
f_{1,1} & \cdots & f_{1,N} \\
\vdots  & \ddots & \vdots \\
f_{N,1} & \cdots & f_{N,N}
\end{pmatrix},
$$
where each line represents a full RIO to an OPQ. Then we can conduct geographical query-item dependent figuring smoothly by using a masked matrix and have the final readout of the score function as follows:
\begin{equation}\label{eq:readout}
  \boldsymbol{s}=\sigma_L(M_f\odot M_s),
\end{equation}
where $\boldsymbol{s}=\{\boldsymbol{s_i}\}_{i=1}^N$ is the combination of scores for all OPQs, $M_s$ is a weighted masked matrix, $\odot$ is a dot product and $\sigma_L$ is a list-level function used to consider inter-item dependencies, which can be an activating function like \textit{softmax} \cite{ai2019learning} or learning function like MLP. 

We can further sort the predicted results and recommend the top $n$ parking spaces. Let $R_{d}^{N}$ be a ranking of OPQ $d$, then we have a $N$-length list of:
\begin{equation}\label{eq:recommend}
R_{d}^{N}=sort\left(F_{d,0},F_{d,1},...,F_{v,n}\right).
\end{equation}

\subsubsection{Objective and Model Training.}

Finally, to implement Equation \ref{L}, the proposed model is trained in a supervised-learning manner with the goal to minimize the distance between lists of relevant labels $\boldsymbol{y}$ and predicted scores $\boldsymbol{\hat{y}}=\boldsymbol{s}$ for citywide OPQ. Recalling that $S_\theta$ assigns a score to each item ({\em i.e.}, parking space), we first simply used a mean square error (MSE) loss function as follows:
\begin{equation}
 l_m(\boldsymbol{y},\boldsymbol{s}) = \sum\limits_{i}{\parallel \boldsymbol{y_i}-\boldsymbol{s_i}\parallel^{2}},
\end{equation}
where $\boldsymbol{y}$ and $\boldsymbol{s}$ indicate full lists of items for label and model output correspondingly. Though we have already considered inter-item dependency in the score function,
we still consider it in the objection here. Specifically, we take advantage of a \textit{Softmax loss} to embed listwise relationship:
\begin{equation}
 l_s(\boldsymbol{y},\boldsymbol{s}) = -\sum\limits_{i}{y_i log\left(\frac{exp\left(s_i\right)}{\sum_j exp\left(s_j\right)}\right)}.
\end{equation}
Finally, the objective function of our proposed model is defined as:
\begin{equation}\label{eq:objective}
 \mathcal{L} = l_m(\boldsymbol{y},\boldsymbol{s}) + \lambda l_s(\boldsymbol{y},\boldsymbol{s}) + \parallel\!\theta\!\parallel_2,
\end{equation}
where $\lambda$ is a hyperparameter, $\theta$ is the model parameter been regularized by L2 and dropout \cite{srivastava2014dropout} is also applied to avoid over-fitting.
\section{Experiments}

\begin{table*}[h]
\caption{Metrics Performance in Hong Kong and San Francisco.}
\centering
\label{tab:modelcompare}
\begin{tabular}{c|cccc|cccc}
\hline
\multirow{2}{*}{Model} & \multicolumn{4}{c|}{Hong Kong}                                                            & \multicolumn{4}{c}{San Francisco}                                                         \\
                       & NDCG@1               & NDCG@5               & MAP@1                & MAP@5                & NDCG@1               & NDCG@5               & MAP@1                & MAP@5                \\ \hline
T-GCN                  & 0.614±0.077          & 0.361±0.048          & 0.186±0.016          & 0.292±0.043          & 0.816±0.090          & 0.480±0.017          & 0.388±0.009          & 0.773±0.052          \\
DCRNN                  & 0.526±0.071          & 0.311±0.046          & 0.157±0.028          & 0.285±0.044          & 0.815±0.090          & 0.475±0.025          & 0.379±0.018          & 0.772±0.052          \\
GAMCN                  & 0.614±0.077          & 0.361±0.048          & 0.186±0.016          & 0.292±0.043          & 0.808±0.086          & 0.432±0.029          & 0.372±0.019          & 0.772±0.052          \\
STPGCN                 & 0.603±0.074          & 0.355±0.046          & 0.183±0.018          & 0.291±0.043          & 0.812±0.092          & 0.478±0.021          & 0.385±0.013          & 0.773±0.051          \\ \hline
T-GCN*                 & 0.707±0.149          & 0.879±0.055          & 0.628±0.132          & 0.822±0.076          & 0.890±0.064          & 0.937±0.040          & 0.945±0.034          & 0.986±0.010          \\
DCRNN*                 & 0.799±0.101          & 0.840±0.069          & 0.582±0.155          & 0.800±0.088          & 0.920±0.050          & 0.949±0.042          & 0.943±0.056          & 0.988±0.018          \\
GAMCN*                 & 0.785±0.117          & 0.876±0.062          & 0.583±0.146          & 0.794±0.087          & 0.916±0.056          & 0.916±0.056          & 0.927±0.058          & 0.982±0.021          \\
STPGCN*                & 0.827±0.097          & 0.900±0.052          & 0.610±0.142          & 0.830±0.074          & 0.922±0.044          & 0.956   ±0.032       & 0.964±0.038          & 0.990±0.014          \\ \hline
listMLE                & 0.487±0.119          & 0.522±0.127          & 0.258±0.210          & 0.515±0.242         & 0.627±0.156          & 0.686±0.106          & 0.585±0.267          & 0.924±0.147          \\
AppNDCG                & 0.486±0.112          & 0.559±0.123          & 0.249±0.184          & 0.491±0.239          & 0.631±0.155          & 0.679±0.105          & 0.586±0.267          & 0.924±0.148          \\
Proposed               & \textbf{0.910±0.086} & \textbf{0.923±0.053} & \textbf{0.782±0.131} & \textbf{0.870±0.065} & \textbf{0.956±0.022} & \textbf{0.966±0.021} & \textbf{0.979±0.015} & \textbf{0.994±0.005} \\ \hline
\end{tabular}
\end{table*}

\begin{table*}[!t]
\caption{Comparison of different model sizes.}
\centering
\label{tab:modelsize}
\begin{tabular}{c|ccccccc}
\hline
\multirow{2}{*}{Data   set} & \multicolumn{7}{c}{Number   of parameters for Models}           \\
                            & T-GCN & DCRNN & GAMCN & STPGCN & listMLE & ApproNDCG & Proposed \\ \hline
Hong   Kong                 & 216K  & 275k  & 13.7M & 381k   & 27k     & 27k       & 11k      \\
San   Francisco             & 221k  & 362k  & 110M  & 393k   & 25k     & 25k       & 70k      \\ \hline
\end{tabular}
\end{table*}

\begin{table*}
\centering
\caption{Comparison of models in different scenarios.}
\label{tab:modelcompare_scenarios}
\begin{tabular}{c|cclcc|ccclc} 	
\cline{1-6}\cline{7-11}
\multirow{2}{*}{{Models}}   & \multicolumn{5}{c|}{{Hong Kong}}                                                                                                                                                                                & \multicolumn{5}{c}{{San Francisco}}                                                                                                                                                                              \\
                                                          & {NDCG@1}               & \multicolumn{2}{c}{{NDCG@5}}               & {MAP@1}                & {MAP@5}                & {NDCG@1}               & {NDCG@5}               & \multicolumn{2}{c}{{MAP@1}}                & {MAP@5}                 \\ 
\cline{1-1}\cline{2-5}\cline{6-6}\cline{7-11}
\multicolumn{11}{c}{{Workdays}}                                                                                                                                                                                                                                                                                                                                                                                                                                                                                                \\ 
\hline
{T-GCN*}                    & {0.815±0.107}          & \multicolumn{2}{c}{{0.876±0.053}}          & {0.616±0.123}          & {0.820±0.069}          & {0.854±0.061}          & {0.913±0.037}          & \multicolumn{2}{c}{{0.926±0.032}}          & {0.982±0.009}           \\
{DCRNN*}                    & {0.790±0.103}          & \multicolumn{2}{c}{{0.832±0.071}}          & {0.577±0.143}          & {0.808±0.084}          & {0.899±0.053}          & {0.933±0.049}          & \multicolumn{2}{c}{{0.927±0.068}}          & {0.983±0.023}           \\
{GAMCN*}                    & {0.776±0.118}          & \multicolumn{2}{c}{{0.873±0.059}}          & {0.583±0.141}          & {0.795±0.081}          & {0.897±0.060}          & {0.925±0.053}          & \multicolumn{2}{c}{{0.909±0.069}}          & {0.977±0.027}           \\
{STPGCN*}                   & {0.826±0.101}          & \multicolumn{2}{c}{{0.897±0.051}}          & {0.617±0.136}          & {0.832±0.072}          & {0.904±0.043}          & {0.945±0.035}          & \multicolumn{2}{c}{{0.956±0.047}}          & {0.989±0.017}           \\
{listMLE}                   & {0.491±0.116}          & \multicolumn{2}{c}{{0.567±0.126}}          & {0.259±0.206}          & {0.520±0.242}          & {0.594±0.146}          & {0.665±0.101}          & \multicolumn{2}{c}{{0.563±0.269}}          & {0.925±0.151}           \\
{AppNDCG}                   & {0.476±0.108}          & \multicolumn{2}{c}{{0.555±0.121}}          & {0.222±0.178}          & {0.494±0.238}          & {0.595±0.146}          & {0.665±0.102}          & \multicolumn{2}{c}{{0.564±0.289}}          & {0.925±0.151}           \\
{Proposed}                  & \textbf{{0.907±0.087}} & \multicolumn{2}{c}{\textbf{{0.922±0.052}}} & \textbf{{0.785±0.137}} & \textbf{{0.868±0.061}} & \textbf{{0.924±0.035}} & \textbf{{0.956±0.021}} & \multicolumn{2}{c}{\textbf{{0.972±0.019}}} & \textbf{{0.993±0.005}}  \\ 
\cline{1-1}\cline{2-5}\cline{6-6}\cline{7-11}
\multicolumn{11}{c}{{Weekends}}                                                                                                                                                                                                                                                                                                                                                                                                                                                                                               \\ 
\hline
{T-GCN*}                    & {0.856±0.092}          & {0.884±0.058}          & \multicolumn{2}{c}{{0.645±0.142}}          & {0.824±0.084}          & {0.936±0.025}          & {0.967±0.013}          & {0.970±0.013}          & \multicolumn{2}{c}{{0.992±0.006}}           \\
{DCRNN*}                    & {0.813±0.097}          & {0.853±0.063}          & \multicolumn{2}{c}{{0.589±0.172}}          & {0.785±0.091}          & {0.949±0.024}          & {0.969±0.017}          & {0.964±0.022}          & \multicolumn{2}{c}{{0.994±0.005}}           \\
{GAMCN*}                    & {0.801±0.113}          & {0.879±0.065}          & \multicolumn{2}{c}{{0.584±0.153}}          & {0.794±0.096}          & {0.942±0.034}          & {0.961±0.022}          & {0.952±0.028}          & \multicolumn{2}{c}{{0.988±0.009}}           \\
{STPGCN*}                   & {0.828±0.092}          & {0.893±0.054}          & \multicolumn{2}{c}{{0.601±0.150}}          & {0.826±0.077}          & {0.948±0.022}          & {0.973±0.014}          & {0.977±0.019}          & \multicolumn{2}{c}{{0.994±0.006}}           \\
{listMLE}                   & {0.481±0.125}          & {0.546±0.128}          & \multicolumn{2}{c}{{0.255±0.219}}          & {0.507±0.241}          & {0.692±0.154}          & {0.728±0.101}          & {0.629±0.258}          & \multicolumn{2}{c}{{0.921±0.139}}           \\
{AppNDCG}                   & {0.471±0.118}          & {0.537±0.125}          & \multicolumn{2}{c}{{0.221±0.197}}          & {0.486±0.241}          & {0.693±0.152}          & {0.731±0.096}          & {0.617±0.259}          & \multicolumn{2}{c}{{0.920±0.141}}           \\
{Proposed}                  & \textbf{{0.916±0.084}} & \textbf{{0.925±0.054}} & \multicolumn{2}{c}{\textbf{{0.779±0.120}}} & \textbf{{0.872±0.072}} & \textbf{{0.960±0.016}} & \textbf{{0.978±0.009}} & \textbf{{0.984±0.009}} & \multicolumn{2}{c}{\textbf{{0.996±0.003}}}  \\ 
\cline{1-1}\cline{2-5}\cline{6-6}\cline{7-11}
\multicolumn{11}{c}{{Daytime}}                                                                                                                                                                                                                                                                                                                                                                                                                                                                                                   \\ 
\hline
{T-GCN*}                    & {0.825±0.101}          & \multicolumn{2}{c}{{0.873±0.042}}          & {0.590±0.134}          & {0.781±0.069}          & {0.867±0.049}          & {0.918±0.032}          & \multicolumn{2}{c}{{0.929±0.027}}          & {0.981±0.007}           \\
{DCRNN*}                    & {0.796±0.099}          & \multicolumn{2}{c}{{0.852±0.055}}          & {0.568±0.145}          & {0.776±0.091}          & {0.909±0.030}          & {0.944±0.023}          & \multicolumn{2}{c}{{0.942±0.028}}          & {0.987±0.011}           \\
{GAMCN*}                    & {0.757±0.108}          & \multicolumn{2}{c}{{0.867±0.046}}          & {0.538±0.137}          & {0.750±0.074}          & {0.891±0.037}          & {0.929±0.028}          & \multicolumn{2}{c}{{0.915±0.035}}          & {0.979±0.014}           \\
{STPGCN*}                   & {0.804±0.090}          & \multicolumn{2}{c}{{0.892±0.040}}          & {0.587±0.134}          & {0.801±0.078}          & {0.906±0.030}          & {0.948±0.020}          & \multicolumn{2}{c}{{0.959±0.020}}          & {0.989±0.008}           \\
{listMLE}                   & {0.530±0.103}          & \multicolumn{2}{c}{{0.623±0.111}}          & {0.261±0.210}          & {0.521±0.241}          & {0.585±0.141}          & {0.674±0.108}          & \multicolumn{2}{c}{{0.559±0.268}}          & {0.925±0.147}           \\
{AppNDCG}                   & {0.508±0.097}          & \multicolumn{2}{c}{{0.606±0.109}}          & {0.223±0.183}          & {0.496±0.239}          & {0.586±0.141}          & {0.675±0.107}          & \multicolumn{2}{c}{{0.556±0.269}}          & {0.925±0.148}           \\
{Proposed}                  & \textbf{{0.912±0.078}} & \multicolumn{2}{c}{\textbf{{0.930±0.038}}} & \textbf{{0.772±0.117}} & \textbf{{0.843±0.063}} & \textbf{{0.928±0.024}} & \textbf{{0.958±0.014}} & \multicolumn{2}{c}{\textbf{{0.972±0.009}}} & \textbf{{0.992±0.003}}  \\ 
\cline{1-1}\cline{2-5}\cline{6-6}\cline{7-11}
\multicolumn{11}{c}{{Nighttime}}                                                                                                                                                                                                                                                                                                                                                                                                                                                                                                 \\ 
\hline
{T-GCN\textsuperscript{*}}  & {0.840±0.101}          & \multicolumn{2}{c}{{0.881±0.062}}          & {0.656±0.122}          & {0.847±0.067}          & {0.903±0.068}          & {0.948±0.040}          & \multicolumn{2}{c}{{0.608±0.262}}          & {0.989±0.009}           \\
{DCRNN\textsuperscript{*}}  & {0.804±0.101}          & \multicolumn{2}{c}{{0.834±0.074}}          & {0.596±0.159}          & {0.813±0.080}          & {0.928±0.058}          & {0.951±0.052}          & \multicolumn{2}{c}{{0.604±0.263}}          & {0.988±0.022}           \\
{GAMCN\textsuperscript{*}}  & {0.805±0.119}          & \multicolumn{2}{c}{{0.879±0.070}}          & {0.619±0.143}          & {0.823±0.083}          & {0.934±0.059}          & {0.948±0.055}          & \multicolumn{2}{c}{{0.981±0.019}}          & {0.983±0.026}           \\
{STPGCN\textsuperscript{*}} & {0.844±0.098}          & \multicolumn{2}{c}{{0.897±0.059}}          & {0.631±0.144}          & {0.847±0.064}          & {0.934±0.044}          & {0.962±0.036}          & \multicolumn{2}{c}{{0.954±0.034}}          & {0.991±0.016}           \\
{listMLE}                   & {0.443±0.117}          & \multicolumn{2}{c}{{0.499±0.113}}          & {0.282±0.246}          & {0.518±0.245}          & {0.663±0.157}          & {0.696±0.101}          & \multicolumn{2}{c}{{0.943±0.070}}          & {0.921±0.146}           \\
{AppNDCG}                   & {0.436±0.112}          & \multicolumn{2}{c}{{0.493±0.112}}          & {0.249±0.226}          & {0.497±0.243}          & {0.664±0.156}          & {0.697±0.099}          & \multicolumn{2}{c}{{0.935±0.071}}          & {0.921±0.147}           \\
{Proposed}                  & \textbf{{0.910±0.090}} & \multicolumn{2}{c}{\textbf{{0.915±0.061}}} & \textbf{{0.789±0.139}} & \textbf{{0.886±0.060}} & \textbf{{0.947±0.037}} & \textbf{{0.970±0.022}} & \multicolumn{2}{c}{\textbf{{0.969±0.048}}} & \textbf{{0.995±0.004}}  \\
\cline{1-1}\cline{2-5}\cline{6-6}\cline{7-11}
\end{tabular}
\end{table*}

In this section, we are mainly interested in the following research questions (RQs):

\begin{itemize}[leftmargin=*]
\item \textbf{(RQ1)} How to compare the proposed method with state-of-the-art models (from both the prediction area and LTR area) in an OPR manner and how well it outperformed others?
\item \textbf{(RQ2)} How does each part (graph and network structures) of our model contribute to the final result and other advantages (like model size) of using these parts? 
\item \textbf{(RQ3)} Practically, how does proposed method perform vs.state-of-the-art baselines in saving citywide on-street parking times?
\end{itemize} 

We thus conduct extensive experiments to answer the above questions.

\subsection{Experimental Setup}
\subsubsection{Dataset.}

To validate the mode, we take advantage of two datasets: Hong Kong Island's on-street parking space data\footnote{\url{https://data.gov.hk/en-data/dataset/hk-td-msd_1-etered-parking-spaces-data}} and on-street parking data in San Francisco\footnote{\url{https://dataverse.harvard.edu/dataset.xhtml?persistentId=doi:10.7910/DVN/YLWCSU}}.

\textbf{Hong Kong.} This dataset is released by the Hong Kong government transport department by recording on-street parking status (vacant or occupied) using a new parking meter every 5 minutes. We use Hong Kong Island's data from Jan 1, 2022, to Apr 30, 2022. After removing bad sensors' data, we finally select 48 location nodes and 28.8k time points.

\textbf{San Francisco.} This dataset contains records of the measured parking availability in San Francisco every 5 minutes from Jun 13, 2013, to Jul 11, 2013. Unlike parking space-based records of Hong Kong, this dataset is at the street level that records capacity and occupied numbers, we treat a ratio of occupied to capacity over $90\%$ as the full-loaded status of the street. After processing, $415$ road nodes and 11.1k time points are acquired.

Both datasets contain detailed location information (longitude and latitude) that is used to build spatial edges between each node, specifically in both data sets, we give an edge between two nodes if the geographic distance between two nodes is less than 50 meters. Then we have 241 edges and 646 edges for Hong Kong and San Francisco correspondingly.

\subsubsection{Baseline Models.} We choose state-of-art models from both prediction and recommendation areas then make some adaptations to OPR that can partly answer \textbf{(RQ1)}. Basically, these baseline models can be categorized into three kinds of tasks: 

\begin{itemize}[leftmargin=*]
\item \textit{Prediction-only task.} This is studied by most researchers. Commonly, they used spatial graphs and temporal sequences to represent the data and then trained a sophisticated model aiming to make more precise predictions. The predicting-based baselines we select are \textbf{T-GCN}\cite{zhao2019t}, \textbf{DCRNN}\cite{li2017diffusion}, \textbf{GAMCN}\cite{qi2022graph}, and \textbf{STPGCN}\cite{9945663}. In this study, predicted results are only used as indicators of future on-street parking availability without recommendation, therefore we can only treat it as owning one candidate in the list. For example, for an OPQ at $d$, those models will come out a list $\{x_d,0,0,...\}$ with $x_d$ indicating the parking state of location $d$ and $0$ is the least value.

\item \textit{Prediction-then-recommendation task.} As we have introduced above, we can make an on-street parking recommendation model directly based on prediction-only models' results. To be specific, we can directly set the score of each item $\{s_i:\forall i\in\mathcal{N}_d\cup \{d\}\}$ using the prediction results, and then the recommendation strategy is the same as our proposed model.
 To distinguish, we add $*$ in the upright of predicting models.

\item \textit{Direct-recommendation task.} We can directly develop a recommendation model for OPR. Though the recommendation models have been widely studied, no attempts have been made in the OPR task. Meanwhile, these models do not handle spatiotemporal data. To compare, we select \textbf{listMLE} and \textbf{ApproxNDCG} as baselines and implement both models using a public source called \textit{allRank}\footnote{\url{https://github.com/allegro/allRank}}. Our proposed OPR-LTR model also belongs to this category. 
\end{itemize}
\subsubsection{Training and Evaluation.} The whole data set is separated into $80\%$, $10\%$, and $10\%$ for training, validating, and testing. We select at most five historic turnover events to make recommendations for future 20  minutes. Our OPR-LTR model is trained by Adam optimizer with a learning rate of 0.002\mbox{\cite{kingma2014adam}}. Additional hyper-parameters are shown in Table~\mbox{\ref{tab:Hyper}} of Appendix A. The training process of the proposed model is shown in Algorithm 3 of Appendix A. Source codes of \textsc{OPR-LTR} are publicly available at \url{https://github.com/HanyuSun/OPR-LTR}. When training prediction-only models, we select the past 12 time points (1 hour), the same as their articles, as the input to predict the parking state of the future 20  minutes. 
To create the ground true ranking list of the parking spaces, the parking lots are ranked based on the proximity of the destination and the duration of the vacancy status. Generally, parking spaces that are nearer and have longer vacancy status will be ranked higher.
The performance of the proposed method is compared with selected baseline models using recommendation-based metrics. As our proposed model is ranking sensitive, we then select Normalised Discounted Cumulative Gain (\textbf{NDCG@n}) and Mean Average Precision (\textbf{MAP@n}). When computing evaluation metrics, the proposed relevant label is used as ground truth for all models. The result can be seen in Table~\ref{tab:modelcompare}.

\subsection{Model Comparison}

As shown in Table \ref{tab:modelcompare}, the comparison results from the experiment are introduced, in which the metric performance at top 1 and 5 are used. We also provide model sizes in Table \ref{tab:modelsize}. By analyzing these results, we answer \textbf{(RQ1)} that is raised at the beginning of this section. 

As indicated in Table~\ref{tab:modelcompare}, our proposed model outperforms all other models in both cities in terms of all the metrics and model sizes, which demonstrates the advantage of the proposed model in handling OPR tasks. Specifically, we notice an NDCG@1 improvement of $73.1\%$ from a prediction-only model and $89.6\%$ from another direct-recommendation method which means both initial feature extraction and later item scoring are important for the final recommendation results, the same conclusion can be gotten from all other metrics. We also find after a recommendation based on the prediction-only model, the results are improved by over $50\%$, it proves the effect of recommendation as well as its basic scoring logic but it also indicates that a two-step recommendation will affect the coherent of the model that will eventually hurt the recommending results. Besides, we find that from NDCG@1 to NDCG@5, prediction-only models like T-GCN perform worse (decrease by $70.1\%$), while direct-recommendation methods like listMLE perform better (increase by $7.2\%$) which has proved again the disadvantage of prediction-only method in an OPR task. We also find that the overall performance of San Francisco is better than Hong Kong. Although the model of San Francisco is larger, the average number of edges is half of Hong Kong (which is 2 to 5), which indicates the dense of parking spots is a big effect on the OPR result. What's more, to demonstrate the robustness of the proposed model, we compare the effectiveness of the proposed model with other methods in different scenarios of workday, weekend, daytime, and nighttime. As shown in Table~\mbox{\uppercase\expandafter{\romannumeral 3}}, the proposed method outperforms all other methods in these four scenarios.

\subsection{Ablation and Sensitivity Analysis}

As discussed in Section~\ref{sec:formulation}, an event-then-graph convolutional layer followed by a learning-based score function is used to aggregate heterogeneous features of ESGraph and readout a ranked OPR list. In this section, to answer \textbf{(RQ2)}, we explain the effects of turnover events and the heterogeneous feature aggregating and updating layer as well as the contribution of each dependency to the performances of the model in aspects of metrics and size. We explain the effects of two main hyper-parameters:  $\alpha$ (number of history turn-over events) and $\beta$ (number of GCN updating layers), as well as the contribution of each dependency to the performances of the model in aspects of metrics and size.

\subsubsection{Effects of ESGraph Representation.}

\begin{figure}
  \centering
  \includegraphics[width=\linewidth]{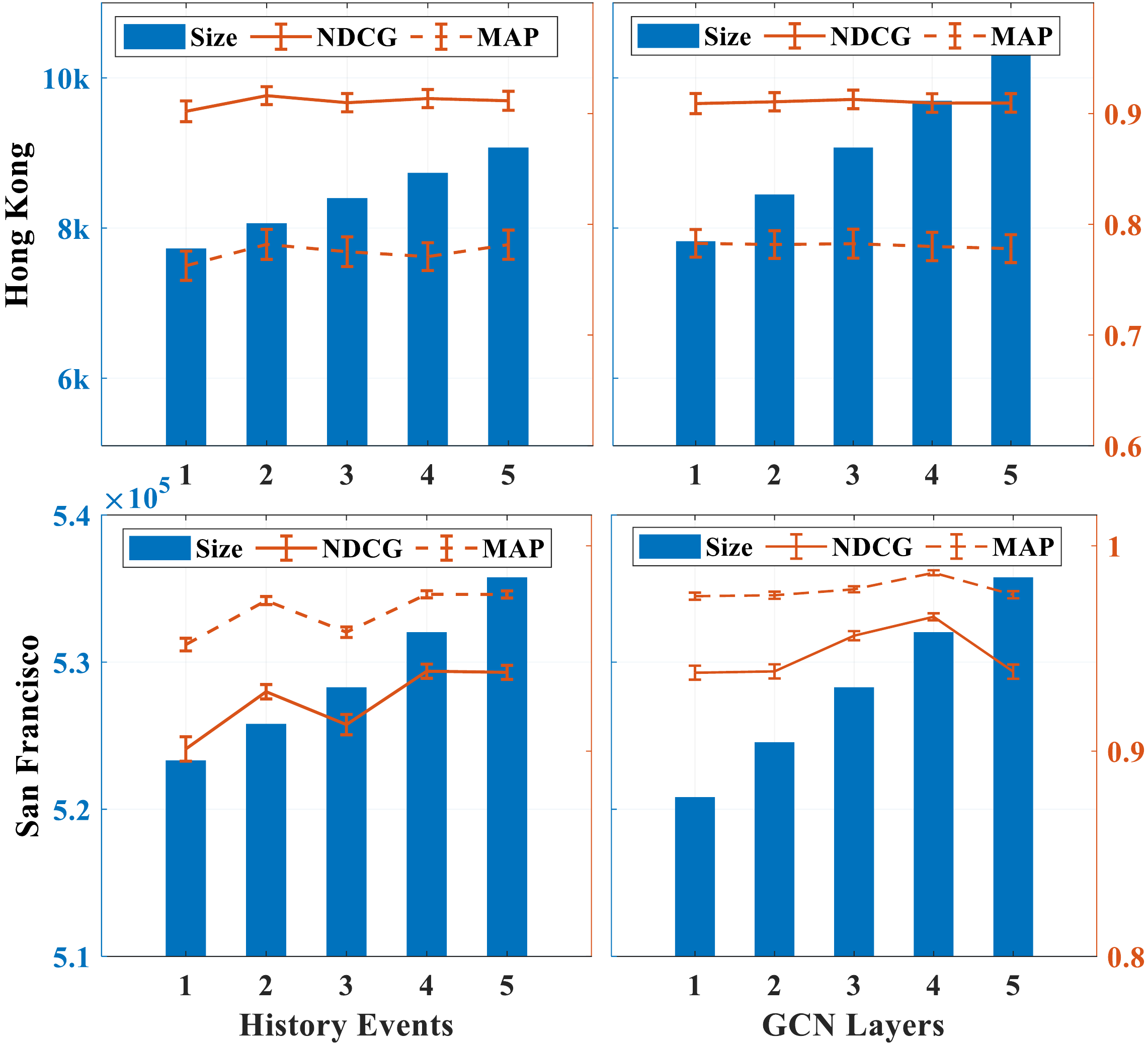}
  \caption{Study of ESGraph Representation. Two hyper-parameters of $\alpha$ (number of history turn-over events) and $\beta$ (number of GCN updating layers) are compared separately to indicate their effect on model performances.}
  \label{ESGraph Representing}
\end{figure}

In model comparison, the proposed model acquires good performances in both recommendation results and model size. Because the ESGraph uses the turnover events as history features can carry more information in each step, it thus can reduce the following representative model many steps to achieve high performance. As shown in Fig.~\ref{ESGraph Representing}, we find that the model has achieved considerably good performances only aggregating several turn-over events of ESGraph ($\alpha=1,2,3$ for Hong Kong, $\alpha=4,5$ for San Francisco) and updating with two to three graph convolutional layers ($\beta=3,4$ for both cities) while usually a STGraph based method needs twelve or more history steps and corresponding updating times. Besides, we find that using more turnover events or updating layers will not increase model size rapidly, this is mainly benefit from the event-then-graph structure, the model will first aggregate turnover events and then combine with the graph but not aggregate the whole graph every time more history steps are required. Theoretically, different spatial vertices in an ESGraph can represent a different number of turnover events without worrying about the inconsistency of aggregation.

\subsubsection{Effect of Interdependency Learning.}
The effect of inter-query-item (Q-Is) and inter-item (I-Is) dependency on the model size and metrics results are discussed here. The raw model is an MLP layer, after which Q-Is are added in an order of spatial Q-Is, temporal Q-Is learning and I-Is learning. When adding I-Is, learning methods with and without MLP along with two different activating methods are considered separately. As shown in Table \ref{tab:dependencycompare}, the first row is an MLP layer, then the contributions of a spatial masked matrix (+Mask), temporal dependency learning (+Temporal), and GLU to ranking results have been revealed; trainable parameters' comparison also proves GLU's contribution in reducing model size, especially in a larger graph. When analyzing the part of I-Is, we also find that the MLP layer has not improved the outcoming results, and the performance of activating functions varies in different data sets.

\begin{table}[]
\caption{Different Dependencies' Contribution Study.}
\label{tab:dependencycompare}
\begin{tabular}{c|c|cc|cc}
\hline
\multirow{2}{*}{Part}                                                      & \multirow{2}{*}{Method} & \multicolumn{2}{c|}{Hong Kong} & \multicolumn{2}{c}{San Franscio} \\
                                                                           &                         & Param.      & NDCG@1           & Param.       & NDCG@1            \\ \hline
Raw                                                                        & MLP                     & 6k          & 0.816±0.126      & 36k          & 0.886±0.059       \\ \hline
\multirow{3}{*}{\begin{tabular}[c]{@{}c@{}}With\\Q-Is\end{tabular}} & + Mask                  & 6k          & 0.847 ±0.130     & 36k          & 0.901 ±0.046      \\
                                                                           & +   Temporal            & 9.3k        & 0.892 ±0.105     & 54k          & 0.873±0.076       \\
                                                                           & + GLU                   & 9k          & 0.904±0.091      & 53k          & 0.924±0.044       \\ \hline
\multirow{4}{*}{\begin{tabular}[c]{@{}c@{}}With\\I-Is\end{tabular}} & MLP+R                   & 11k         & 0.909±0.085      & 71k          & 0.930±0.035       \\
                                                                           & MLP+S                   & 11k         & 0.870±0.124      & 71k          & 0.935 ±0.038      \\
                                                                           & Relu(R)                 & 11k         & 0.912±0.086      & 54k          & 0.940 ±0.034      \\
                                                                           & Softmax(S)              & 9.3k        & 0.880 ±0.122     & 54k          & 0.956±0.022       \\ \hline
\end{tabular}
\end{table}

\subsection{Practical Effectiveness}

\begin{figure}
  \centering
  \includegraphics[width=0.8\linewidth]{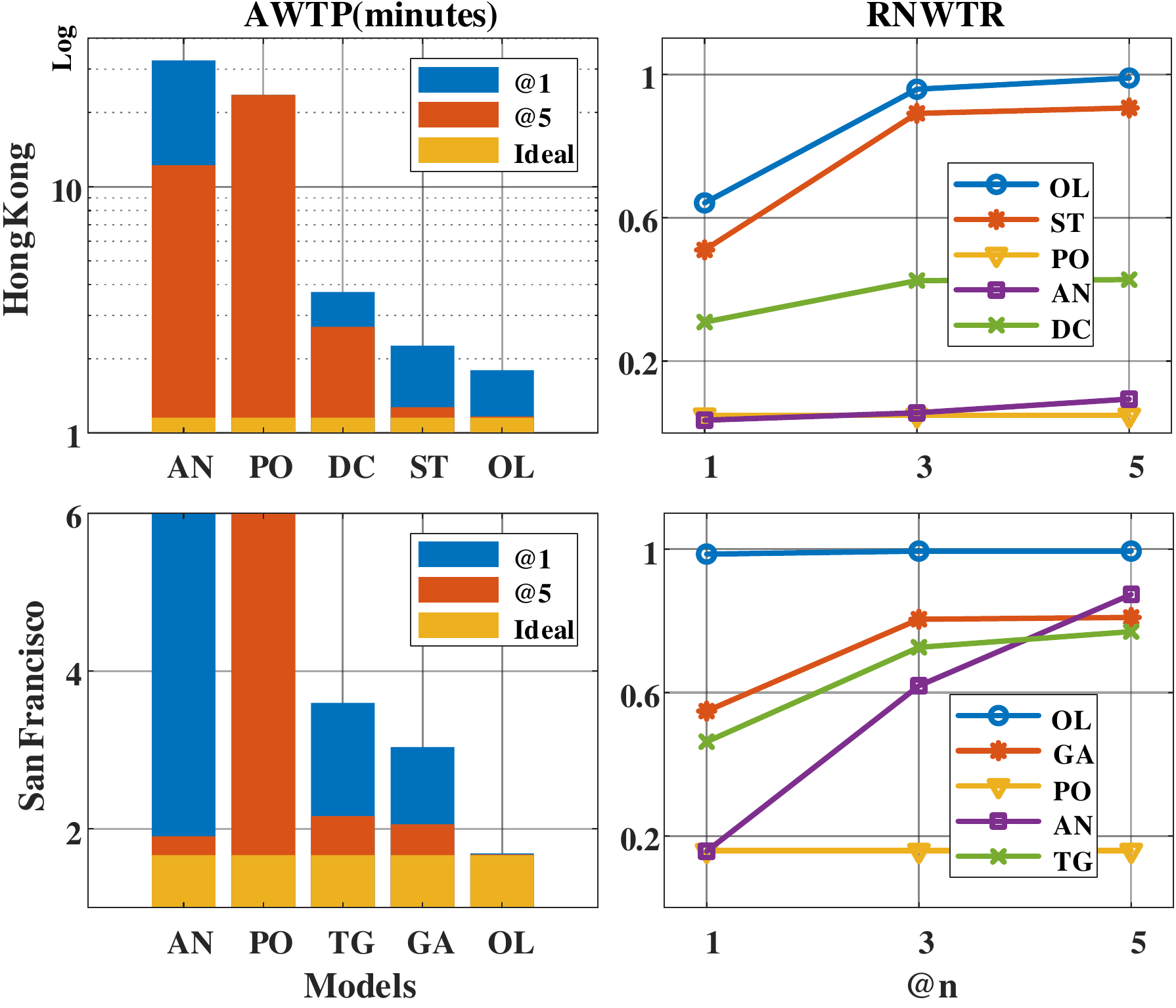}
  \caption{Study of Practical Effectiveness. Compared models are AppNDCG (AN), Prediction-Only models (PO), T-GCN (TG), DCRNN (DC), GAMCN (GA), STPGCN (ST), and the Proposed OPR-LTR (OL).}
  \label{Study of Practical Effectiveness}
\end{figure}

We analyze the practical effectiveness \textbf{(RQ3)} of our proposed model in saving drivers'  cruising time. The parking process after a driver acquired the recommended list can be seen as driving from the top first to top end parking location until a vacant space is found, while all recommended locations are neighbors with a specific location, no extra time will be wasted as long as one of recommended location is satisfied. To this end, we first design two unique metrics to measure the usage and efficiency at a citywide level: Absolute Waiting Time in Parking (AWTP) and Relative Non-waiting Time Ratio (RNWTR):
\begin{equation}
AWTP@n = \frac{1}{N}\sum\limits_{i=1}^{N}{minimum\{\mathcal{T} (R_{i}^{n})\}},
\end{equation}
where $N$ indicates all possible OPQs, $\mathcal{T}$ computes the real waiting time of each recommendation. Let IAWTP@n equal the ideal minimum waiting time, then we have: 
\begin{equation}
RNWTR@n = \frac{IAWTP@n}{AWTP@n}.
\end{equation}
The comparing results are shown in Fig.~\ref{Study of Practical Effectiveness}, from which we find the proposed OPR-LTR model (short for PR) has largely decreased the waiting time and increased the parking efficiency of on-street parking.

\section{Conclusion}

This paper develops a practical task to directly recommend a parking space to drivers given a specific query, and we have demonstrated that such a task has great potential in saving drivers' on-street searching time in a citywide manner, compared to the prediction-only and prediction-then-recommendation models.
In terms of model development, we highlight the importance of turnover events in parking recommendation, and hence an ESGraph-based data representation and an event-then-graph-based convolution network are developed. The ESGraph is proven to have better representation power and lower space/time complexity than the STGraph.
Numerical experiments also demonstrate the outperformance of the proposed model, and especially, the proposed model could at most reduce the time in cruising for parking by $95.07\%$ in Hong Kong and $84.02\%$ in San Francisco. 
Future studies will focus on designing a real-time parking recommendation system that can be used in the real world. Parking competitions can also be considered when providing the recommended parking lots. Additionally, ESGraph can be applied to other applications, such as signal timing, system failure, etc, and we plan to develop more generalized heterogeneous graph networks for smart civil infrastructure systems.

\appendix
\subsection{Implementation Details}
The pseudocode for the edge contraction is shown in Algorithm~\ref{alg:cap}, the ESGraph Merging algorithm is presented in Algorithm~\ref{alg:ESGraph}, and the training process of OPR\_LTR is shown in Algorithm~\ref{alg:OPR-LTR}. 

\renewcommand{\algorithmicrequire}{\textbf{Input:}}  
\renewcommand{\algorithmicensure}{\textbf{Output:}} 
\begin{algorithm}[h]
\caption{Edge contraction for a Temporal Path: $\Gamma$}\label{alg:cap}
\begin{algorithmic}
\Require Temporal path $(V_t, E_t)$, node attributes $X_v\in\{0,1\} (\forall v\in V_t$), $V_t=\{n_1, n_2, ..., n_T\}$, $E_t=\{(n_i, n_{i+1}), i=1,...,T-1\}$, weight set $W$, counter $c$ for $(V_t, E_t)$, counter $c^{\prime}$ for  output , weight $w$
\Ensure $(U, \mathcal{E}, W_\varepsilon)$ 
\State $u_1 \gets v_1$ 
\State $w \gets 1$
\State $c \gets 0$
\State $c^{\prime} \gets 0$
\While{$c < T$}
\State $c \gets c+1$
\If{$X_{v_c} \neq X_{n_{c+1}}$}
    \State $\mathcal{E} \gets E_t\cup \{(n_c, n_{c^{\prime}})\}$
    \State $W_{c^{\prime}} \gets w$ 
    \State $w \gets 1$
\Else
    \State $E_t \gets E_t\backslash\{(n_c, n_{c+1}\}$ 
    \State $V_t \gets V_t\backslash\{n_c, n_{c+1}\}$
    \State $w \gets w+1$
\EndIf
\If{$c=T-1$}
    \State $\mathcal{E} \gets E_t\cup \{(n_{c^{\prime}}, n_{c+1})\}$
\EndIf
\EndWhile
\State $U \gets V_t$
\State $W_\varepsilon \gets W$
\end{algorithmic}
\end{algorithm}

\begin{algorithm}[h]
\caption{ESGraph Merging Algorithm: $M$}\label{alg:ESGraph}
\begin{algorithmic}
\Require Location set $L$, Length of time $T$, History attribute of each location $X$, index $i$
\Ensure $(V, E, U, \mathcal{E}, W_\varepsilon)$
\State $V \gets \text{ENUMERATE}(L)$
\State $E \gets \text{ADJACENT}(L,V)$
\State $i \gets 0$
\While{$i<\parallel\! V\! \parallel$}
\State $V_{i,T} \gets \{x_{i,1},x_{i,2},...,x_{i,T}\}$
\State $E_{i,T} \gets \{(n_{i,j},n_{i,j+1}):\forall n_j\in V_{i,T}\}$
\State $(U_i, \mathcal{E}_i, W_{\varepsilon_i}) \gets \Gamma(V_{i,T},E_{i,T})$
\State $U\gets U \cup U_i$
\State $\mathcal{E} \gets \mathcal{E} \cup \mathcal{E}_i$
\State $W_\varepsilon \gets W_\varepsilon\cup W_{\varepsilon_i}$
\EndWhile
\end{algorithmic}
\end{algorithm}

\begin{algorithm}[h]
\caption{{Training Algorithm for OPR-LTR}\label{alg:OPR-LTR}}
\begin{algorithmic}
\Require {ESGraph-based OPR data set $\mathcal{D}_{ES}$, Number of history events $n$, Start point of training $T_0$, Length of training dataset $L_{train}$, Number of graph convolution-based updating $k$, Initialized the model parameter $\Theta$	}
\Ensure {Trained OPR-LTR mode}
\State {$//$ Training data preparation}
\State {$\mathcal{D} \gets \emptyset$}
\For{{$t=T_0,T_0+1,...,T_0+L_{train}-1$}}
    \State {Training input $\chi \gets \lVert\mathcal{D}_{ES} \rVert_{t}^{n}$}
    \State {Training label $\psi \gets Sort \& List(\lVert\mathcal{D}_{ES} \rVert_{t}^{1})$}
    \State {$\mathcal{D} \gets \{tanh(\chi),minmax(\psi)\} \cup \mathcal{D}$}
\EndFor
    \State {$\mathcal{D} \gets \text{SHUFFLE}(\mathcal{D})$}
\State {$//$ Model training}
\State {$i\gets 1$}
\While{{not stop criteria}}
    \For{{$i<k+1$}}
        \State {$H_{\Theta}^{(i)} \gets \text{AGGREGATE}(\chi)$ based on Eqn.~\ref{eq:aggregate}}
        \State {$Z_{\Theta}^{(i)} \gets \text{UPDATE}(H_{\Theta})$ based on Eqn.~\ref{eq:update}}
        \State {$i \gets i+1$}
    \EndFor
    \State {$\varsigma \gets \text{READOUT}(Z_{\Theta}^{(k)},\chi)$ based on Eqn.~\ref{eq:readout}}
    \State {Update $\Theta$ by minimizing the Eqn.~\ref{eq:objective}}
\EndWhile
\end{algorithmic}
\end{algorithm}

\begin{table}[]
\centering
\caption{Hyper-parameters in Training}
\label{tab:Hyper}
\begin{tabular}{c|cc}

\hline
\multirow{2}{*}{{\textbf{Hyper-parameters}}} & \multicolumn{2}{c}{{\textbf{Values}}}                  \\
                                  & {Hong   Kong}                & {San Francisco} \\ \hline
{Learning   rate}                   & \multicolumn{1}{c}{{0.002}} & {0.002}         \\ 
{Iteration}                         & \multicolumn{1}{c}{{1000}}  & {500}           \\ 
{Batch size}                        & \multicolumn{1}{c}{{128}}   & {128}           \\ 
{$\alpha$}                                 & \multicolumn{1}{c}{{2}}     & {5}             \\ 
{$\beta$}                                 & \multicolumn{1}{c}{{3}}     & {4}             \\ 
{Activating   function}             & \multicolumn{1}{c}{{Relu}}  & {Logistic}       \\ \hline
\end{tabular}
\end{table}

\subsection{Complexity Comparison}
This section discusses the advantages of the proposed ESGraph over STGraph in terms of space and time complexity.
\subsubsection{Space Complexity}
When comparing the space complexity of STGraph and ESGraph, we compute the space occupation  with the growth of data size by using citywide on-street parking data in Hong Kong. Assuming there are $m$ locations and $n$ timepoints recorded in the dataset and using $f_{ST}$ and $f_{ES}$ to represent the space (time) occupation of STGraph and ESGraph.

For STGraph 
$f_{ST}=mn$,
while $f_{ST}$ is fixed for a given $m$ and $n$, we then have:
\begin{equation}
f_{ST} \in O(n^2)\ as\ m \to \infty, n \to \infty.
\end{equation}

For ESGraph, the space occupation is changing with event path. Using $g$ to calculate the space occupation of each path:
\begin{equation}
f_{ES}=\sum\limits_{i=1}^m{g(i)}, \forall g(i)\in [1,2,...,n].
\end{equation}
If  $\forall g(i)=1$, we have the best situation:
\begin{equation}
f_{ES}=m\in O(n)\ as\ m \to \infty.
\end{equation}
If  $\forall g(i)=n$, we have the worst situation:
\begin{equation}
f_{ES}=mn\in O(n^2)\ as\ m \to \infty, n \to \infty.
\end{equation}
Using an aggregate analysis and letting upper bound $T(n)=O(n^2)$, then the amortized cost of ESGraph's space complexity is $T(n)/n = O(n)$. In Fig.~\ref{complexity}, space complexity from five arbitrary 1-hour time intervals using both STGraph and ESGraph have been compared and ESGraph requires less space.
\subsubsection{Time Complexity}
When analyzing time complexity, we first compare the corresponding steps ESGraph $(n_e)$ and STGraph $(n_s)$ will take to conduct the following two tasks:

{\bfseries Task 1: Reviewing states in $n$ time length.} This task is very much like the above space complexity analysis in which $n_s=n$ for each location makes it a complexity of $O(n^2)$. Similarly, $n_e=[1,2,...,n]$ for each location makes it an amortized complexity of $O(n)$. 

{\bfseries Task 2: Reviewing last $l$ turn over events.} For ESGraph, the time complexity is $O(n)$.
For STGraph, $n_s = \sum_{i=1}^l{n_{s,i}}$, where $n_{s,i}\in \mathbb{Z}^+$ is the number of steps for each event reviewing, which is not fixed. If for each location, $n_s=1$, then getting the best situation:
\begin{equation}
f_{ST}=ml, \quad f_{ST} \in O(n)\ as\ m \to \infty,
\end{equation}
which is of little possibility for STGraph.

\begin{figure}[h]
  \centering
  \includegraphics[width=0.9\linewidth]{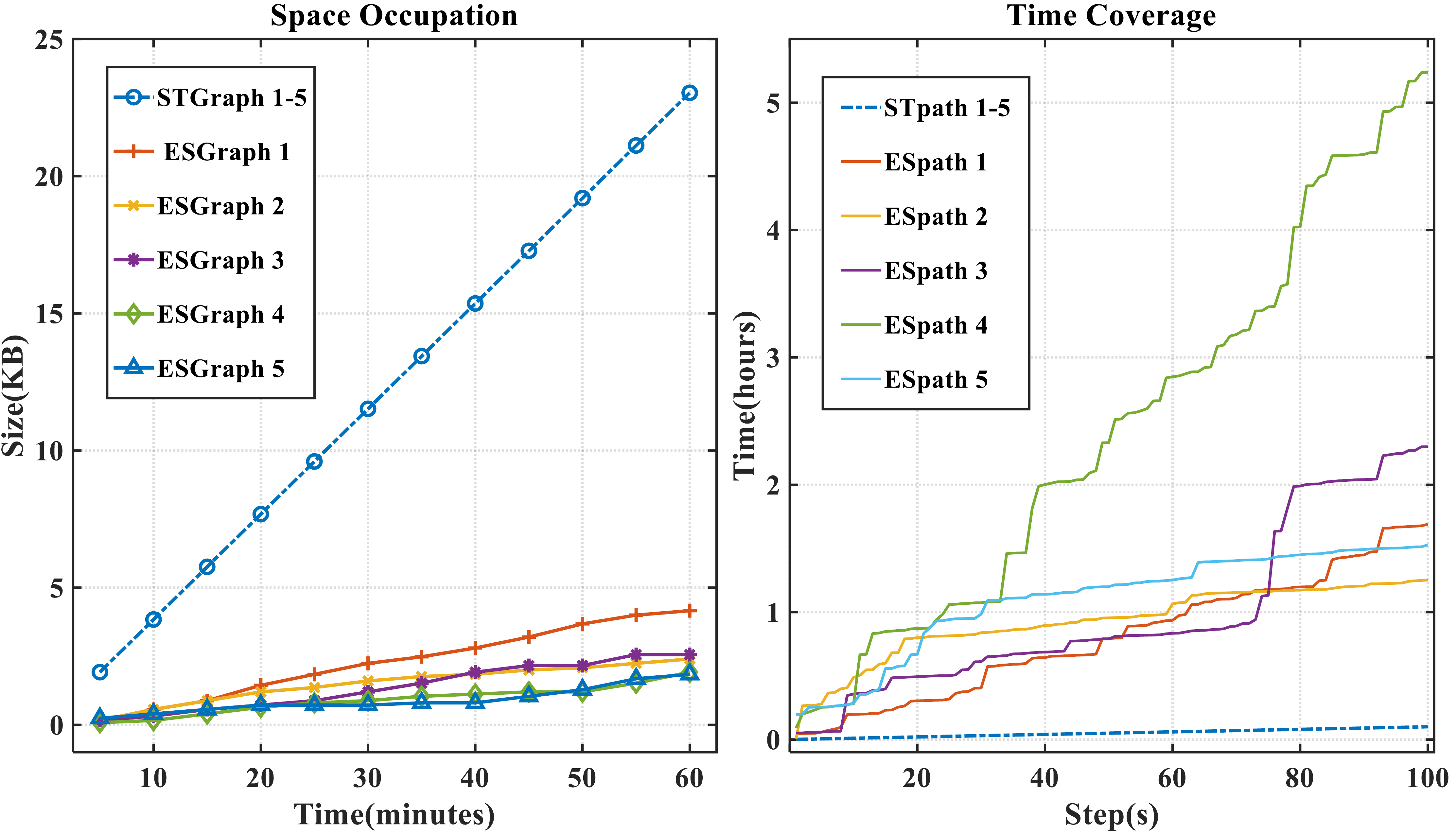}
  \caption{Comparison of the space and time complexity.}
  \label{complexity}
\end{figure}

In ESGraph, the weighted edge after path contracting can store more information than STGraph. As shown in Fig.~\ref{complexity}, when reviewing 100 steps from five arbitrary locations using both graphs, the path of ESGraph (ESpath) can express more time than STGraph (STpath). 

\bibliographystyle{IEEEtran}
\bibliography{ref}

\begin{IEEEbiography}
[{\includegraphics[width=1in,height=1.25in,clip,keepaspectratio]{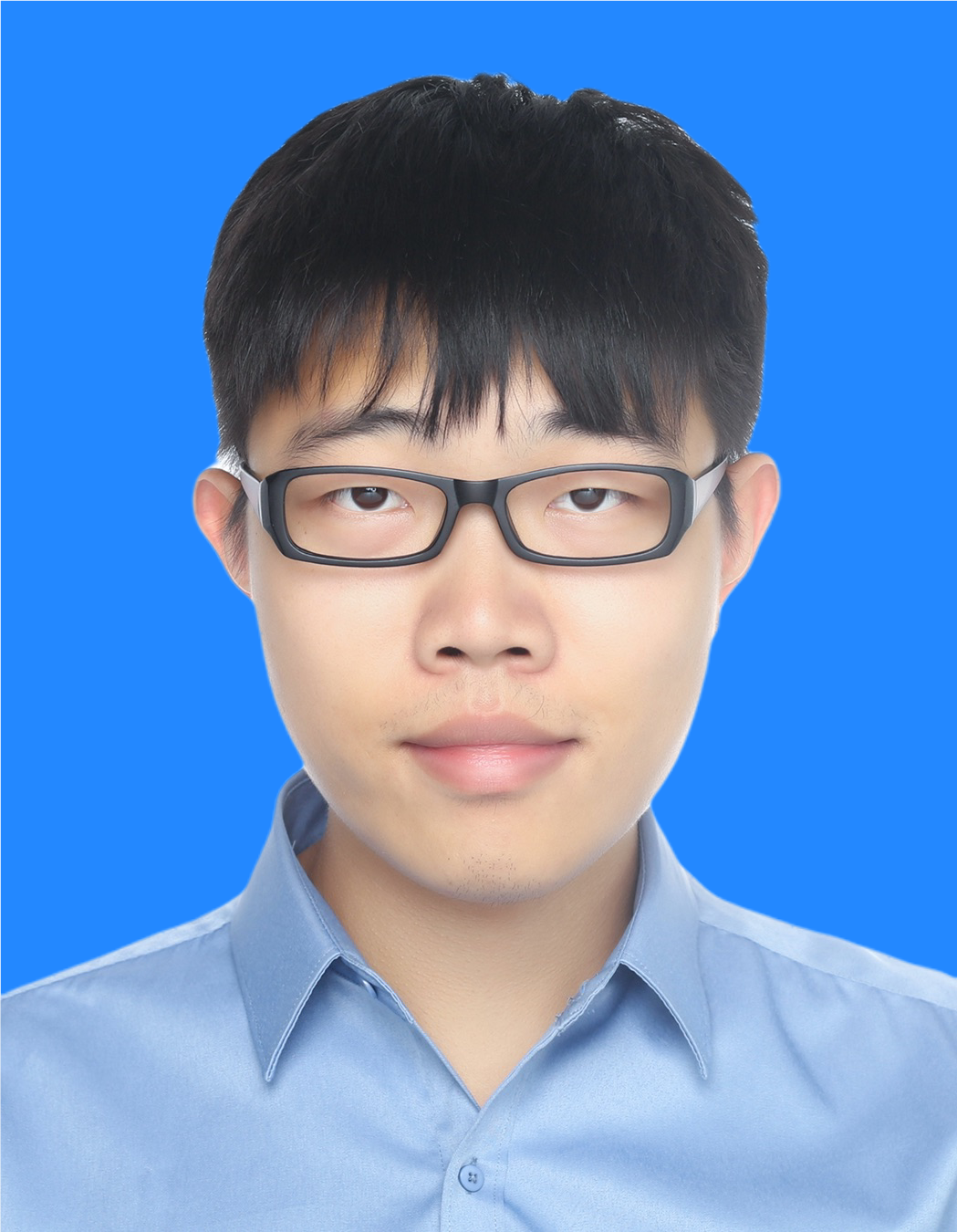}}]{Hanyu Sun} graduated from Beihang University, and he is currently a Research Assistant with the Department of Civil and Environmental Engineering at the Hong Kong Polytechnic University (PolyU). His research interests include intelligent transportation systems (ITS), deep learning, graph mining, and signal processing.
\end{IEEEbiography}

\begin{IEEEbiography}
[{\includegraphics[width=1in,height=1.25in,clip,keepaspectratio]{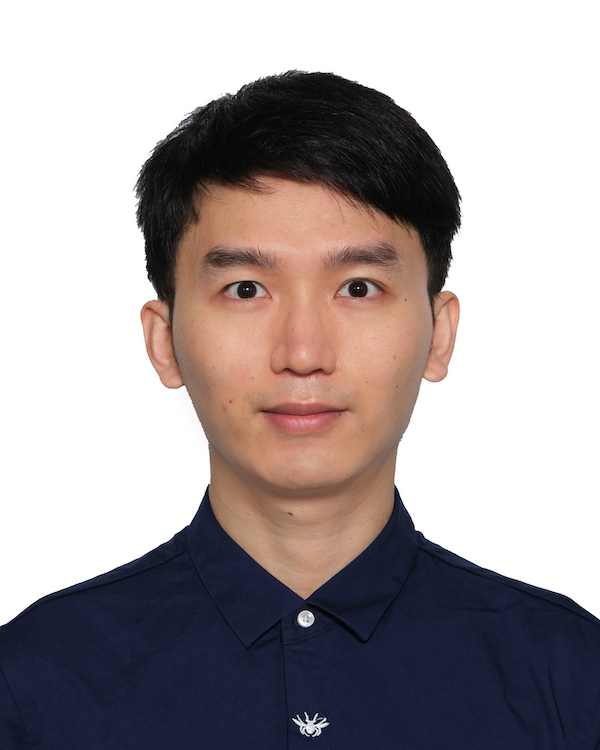}}]{Xiao Huang} is an assistant professor at the Department of Computing, The Hong Kong Polytechnic University. He received Ph.D. in Computer Engineering from Texas A\&M University in 2020, and B.S. in Engineering from Shanghai Jiao Tong University in 2012. Before joining PolyU, he worked as a research intern at Microsoft Research and Baidu USA. He is a program committee member of NeurIPS 2021-2023, AAAI 2021-2023, ICLR 2022-2023, KDD 2019-2023, TheWebConf 2022-2023, ICML 2021-2023, IJCAI 2020-2023, CIKM 2019-2022, WSDM 2021-2023, SDM 2022, and ICKG 2020-2021. He has published over forty papers in top conferences/journals.

\end{IEEEbiography}

\begin{IEEEbiography}[{\includegraphics[width=1in,height=1.25in,clip,keepaspectratio]{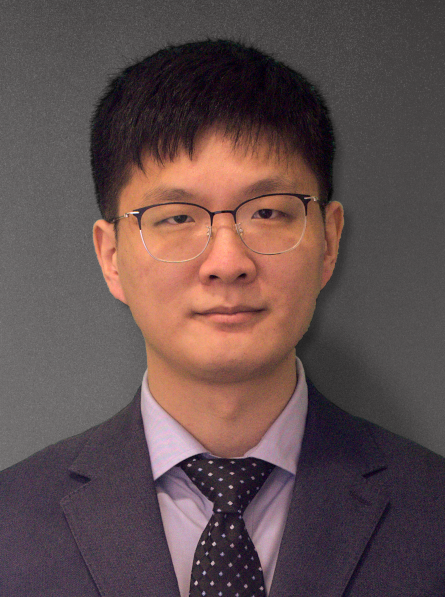}}]{Wei Ma} (IEEE member) received bachelor’s degrees in Civil Engineering and Mathematics from Tsinghua University, China, master degrees in Machine Learning and Civil and Environmental Engineering, and PhD degree in Civil and Environmental Engineering from Carnegie Mellon University, USA. He is currently an assistant professor with the Department of Civil and Environmental Engineering at the Hong Kong Polytechnic University (PolyU). His research focuses on intersection of machine learning, data mining, and transportation network modeling, with applications for smart and sustainable mobility systems. He has received 2020 Mao Yisheng Outstanding Dissertation Award, Best Paper Award (theoretical track) at INFORMS Data Mining and Decision Analytics Workshop, and 2022 Kikuchi Karlaftis Best Paper Award of the TRB Committee on Artificial Intelligence and Advanced Computing Applications.
\end{IEEEbiography}

\end{document}